\documentclass[sigconf]{acmart}

\AtBeginDocument{%
  \providecommand\BibTeX{{%
    \normalfont B\kern-0.5em{\scshape i\kern-0.25em b}\kern-0.8em\TeX}}}

\usepackage{multirow}

\newcommand{\tabincell}[2]{\begin{tabular}{@{}#1@{}}#2\end{tabular}}

\copyrightyear{2021}
\acmYear{2021}
\setcopyright{acmcopyright}\acmConference[CIKM '21]{Proceedings of the 30th ACM International Conference on Information and Knowledge Management}{November 1--5, 2021}{Virtual Event, QLD, Australia}
\acmBooktitle{Proceedings of the 30th ACM International Conference on Information and Knowledge Management (CIKM '21), November 1--5, 2021, Virtual Event, QLD, Australia}
\acmPrice{15.00}
\acmDOI{10.1145/3459637.3482270}
\acmISBN{978-1-4503-8446-9/21/11}

\begin{document}
\fancyhead{}
\title{Learning Joint Embedding with Modality Alignments for 
Cross-Modal Retrieval of Recipes and Food Images}

\author{Zhongwei Xie$^{+,*}$, Ling Liu$^{+}$, Lin Li$^{*}$, Luo Zhong$^{*}$}
\affiliation{%
  \institution{$^{+}$School of Computer Science, Georgia Institute of Technology, Atlanta, Georgia, USA \\
$^{*}$School of Computer Science and Technology, Wuhan University of Technology, Wuhan, Hubei, China}
  \country{}}
  \email{zhongweixie@gatech.edu, lingliu@cc.gatech.edu, {cathylilin, zhongluo}@whut.edu.cn}

\renewcommand{\shortauthors}{Trovato and Tobin, et al.}

\begin{abstract}
This paper presents a three-tier modality alignment approach to learning text-image joint embedding, coined as JEMA, for cross-modal retrieval of cooking recipes and food images. The first tier improves recipe text embedding by optimizing the LSTM networks with term extraction and ranking enhanced sequence patterns, and optimizes the image embedding by combining the ResNeXt-101 image encoder with the category embedding using wideResNet-50 with word2vec. The second tier modality alignment optimizes the textual-visual joint embedding loss function using a double batch-hard triplet loss with soft-margin optimization. The third modality alignment incorporates two types of cross-modality alignments as the auxiliary loss regularizations to further reduce the alignment errors in the joint learning of the two modality-specific embedding functions. The category-based cross-modal alignment aims to align the image category with the recipe category as a loss regularization to the joint embedding. The cross-modal discriminator-based alignment aims to add the visual-textual embedding distribution alignment to further regularize the joint embedding loss. Extensive experiments with the one-million recipes benchmark dataset Recipe1M demonstrate that the proposed JEMA approach outperforms the state-of-the-art cross-modal embedding methods for both image-to-recipe and recipe-to-image retrievals.
\end{abstract}



\maketitle

\section{ Introduction}

\begin{figure}[t]
  \centering
  \includegraphics[scale=0.2]{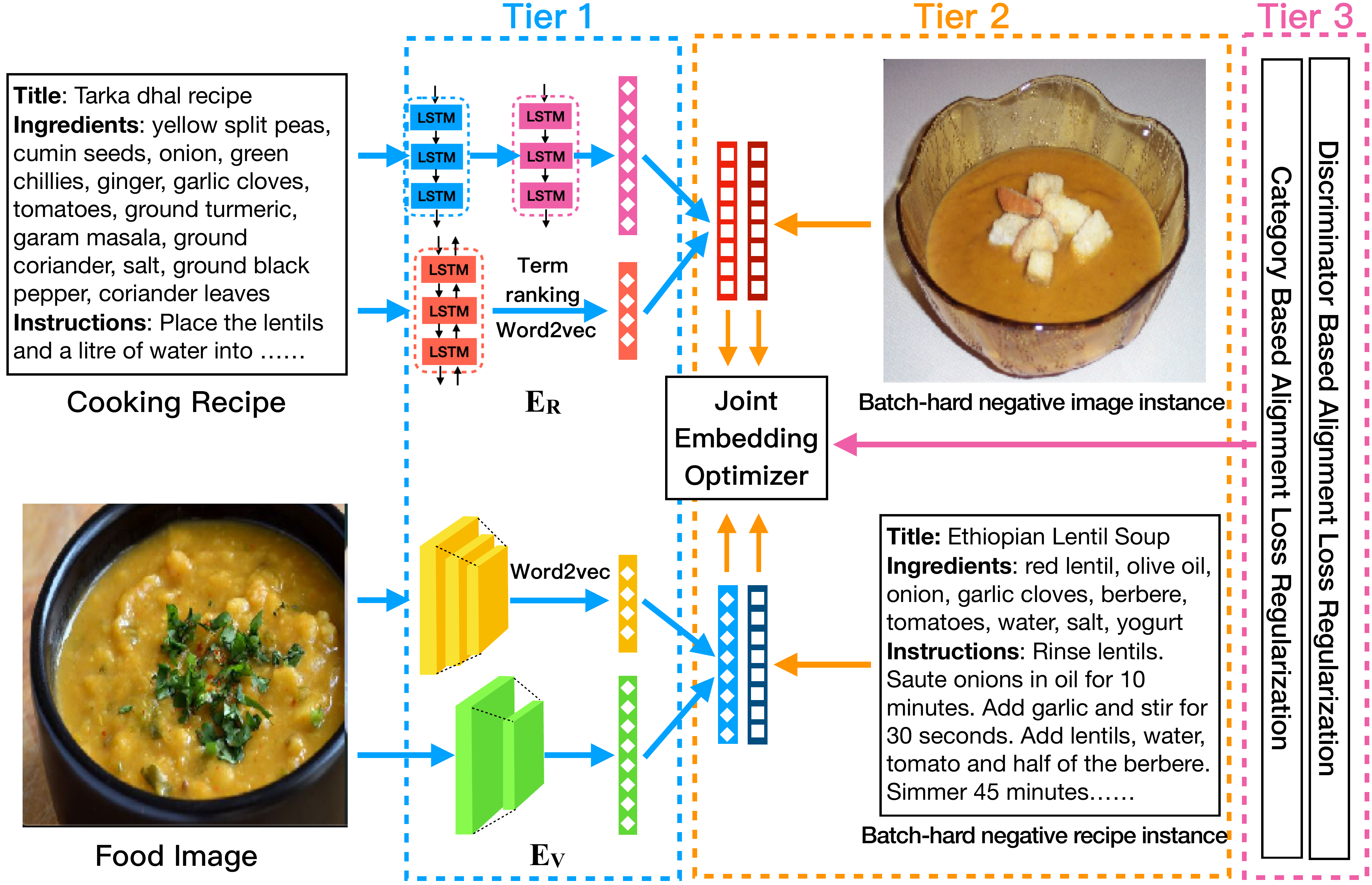}
    \vspace{-0.3cm}
  \caption{{\footnotesize Cross-modal embedding with three-tier modality alignments.}}                    
  \label{simple_framework}
  \vspace{-0.4cm}
\end{figure}

\noindent The prevalence of social media has enabled the rapid growth of  
user-generated online recipes and food images~\cite{Tratter+WWW2017}, such as ``Food.com'' and ``CookEatShare.com''. Most recipes provide ingredients with their quantities and cooking instructions on how ingredients are prepared and cooked (e.g., steamed or deep-fried), providing a new source of references for food intake tracking and health monitoring and management. 
Learning cross-modal joint embeddings has been a growing area of interest for performing image to recipe and recipe to image retrieval tasks. Food images are diverse in terms of background, ingredient composition, visual appearance and ambiguity. Early works~\cite{Jeno+SIGIR-2003,Sun-2011} circumvented this problem by annotating images to perceive their latent semantics.  
These approaches, however, are supervised and require users to annotate at least a small portion of images. In comparison, the unsupervised solutions map images and recipe texts into a shared latent space in which their feature vectors can be compared in terms of vector similarity. Recent approaches have employed deep neural networks to cross-modal retrieval tasks, such as DeViSE\cite{Frome+2013},  
correspondence auto-encoder~\cite{Feng+MM2014}, adversarial cross-modal retrieval~\cite{Wang-2017}, or combing LSTM networks with CNN networks to generate the recipe embedding and image embedding respectively~\cite{Salvador+CVPR2017_JESR, Carvalho+SIGIR2018_AdaMine, JinJinChen+MM2018_AMSR,Hao+CVPR2019_ACME,zhu2019r2gan,lien2020recipe,fu2020mcen}. However, few existing approaches have provided modality alignment optimizations at all three tiers of a cross-modal joint embedding learning process: (1) modality-specific embeddings, such as recipe text embedding and food image embedding; (2) distance-based joint embedding loss optimization; and (3) loss regularization to further optimize cross-modal alignment. 

\begin{figure*}[t] 
  \centering 
  \includegraphics[scale=0.24]{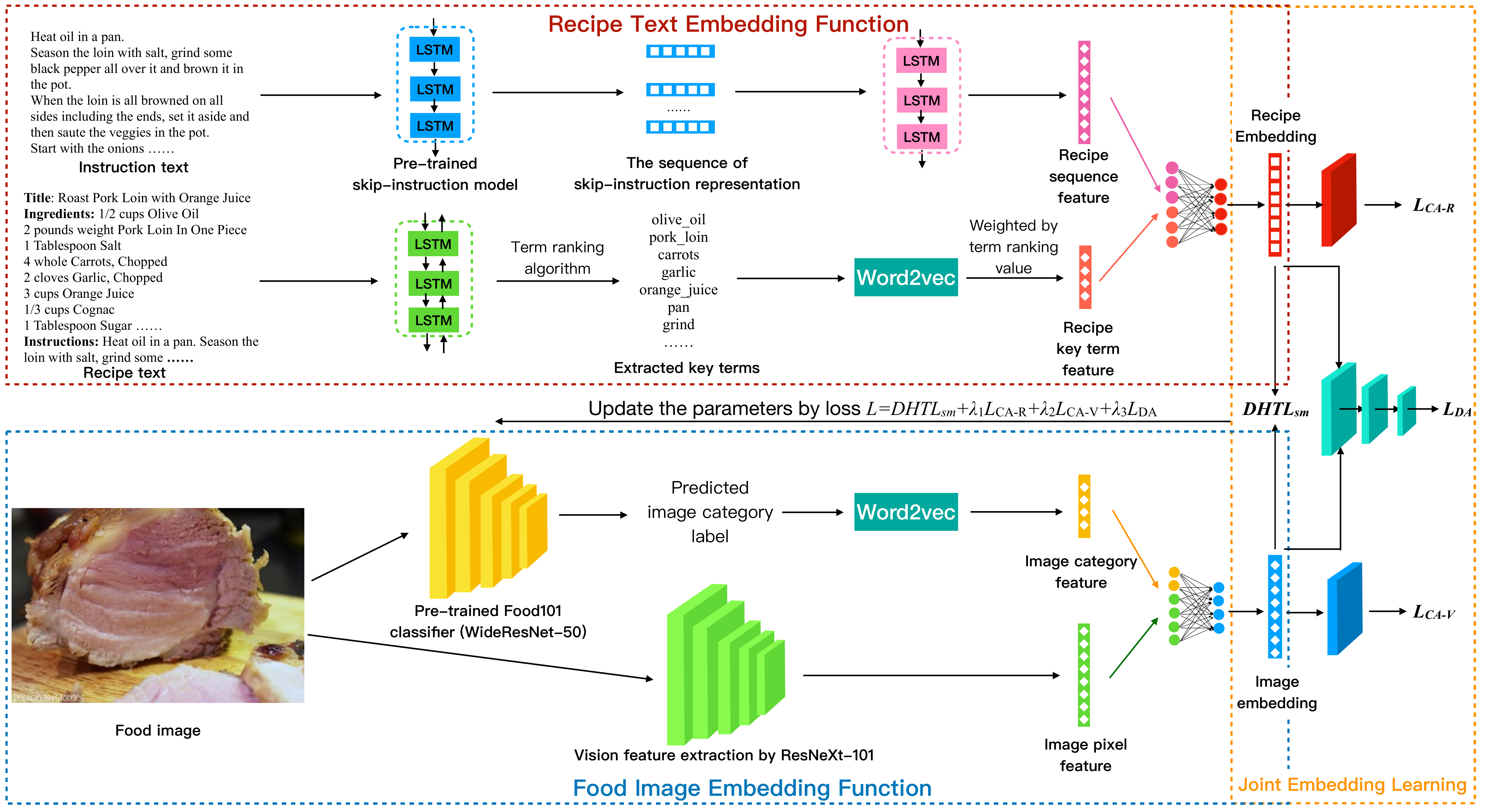}  
  \caption{The system architecture of our JEMA cross-modal embedding learning approach.}
    \vspace{-0.3cm}
  \label{general_framework} 
  \vspace{-0.2cm}
\end{figure*}  

In this paper, we present a three-tier modality alignment optimization framework for learning cross-modal joint embedding, coined as JEMA, aiming for supporting high-performance cross-modal information retrieval tasks. Figure~\ref{simple_framework} provides an illustrative overview of our JEMA three-tier modality alignment optimization process. 
This paper makes three main contributions. {\em First}, we introduce the term extraction and ranking enhanced recipe embedding approach to make the textual embedding more aligned with the visual features captured in the food image embedding, and we enhance image embedding learning with the category information to improve the alignment of the visual information with key ingredients in the recipe. 
{\em Second}, we develop an instance-class double sampling based batch-hard triplet loss with soft-margin optimization, coined as $DHTL_{sm}$, which aims to effectively optimize the textual-visual distance alignment in the joint embedding space. {\em Third}, we further optimize the distance-based $DHTL_{sm}$ cross-modal alignment loss by two types of loss regularizations. The category-based loss regularization employs category-based modality alignment on both recipe and image embeddings to further reduce errors in $DHTL_{sm}$. The cross-modal discriminator-based loss regularization adds the distribution-based alignment between the image and its corresponding recipe as another regularization to $DHTL_{sm}$.  
Extensive experiments are conducted for image-to-recipe and recipe-to-image cross-modal retrieval tasks on the Recipe1M benchmark dataset~\cite{Salvador+CVPR2017_JESR}, consisting of over 800K recipes (title, list of ingredients and cooking instructions) and 1 million associated food images. The evaluation results show that empowered by the three-tier recipe-image alignment optimization techniques, our JEMA approach outperforms existing representative methods in terms of cross-modal retrieval performance for both image-to-recipe and recipe-to-image queries.

 \begin{figure*}[tbp]
  \centering
  \includegraphics[scale=0.42]{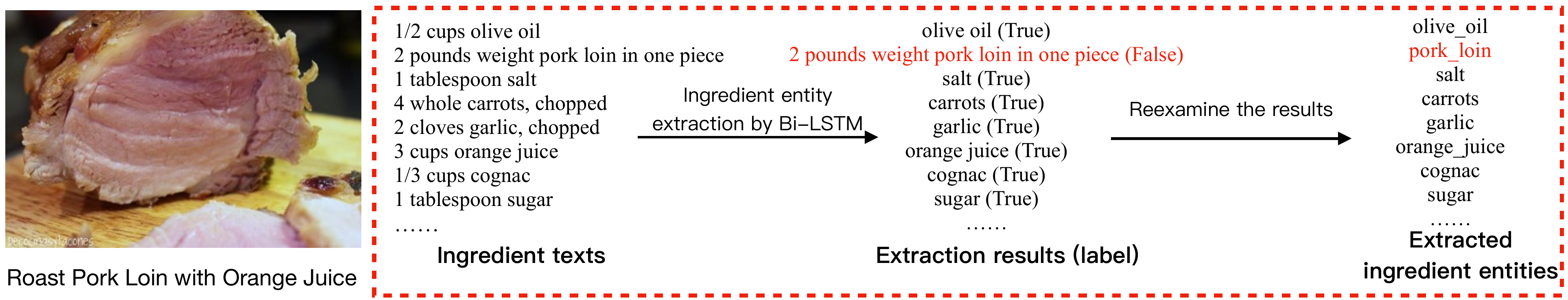}
    \vspace{-0.2cm} 
  \caption{{\small The workflow of extracting the ingredient entities is depicted in the red dashed box using an example recipe and the matched food image. JEMA refines the LSTM extracted key ingredient results in this example: identify the key term ``pork\_loin'' from the false results as the key ingredient, effectively correcting the errors made from LSTM sequence embedding.}} 
  \label{miss-ingr} 
    \vspace{-0.2cm}
\end{figure*}

\section{ Related Work}  
\noindent There are three popular learning tasks on food datasets: (1) {\bf Food recognition} from images, which evolves from kernel-based model~\cite{Joutou+ICIP2009} to DNN approaches~\cite{Kawano+2014,Yanai+2015}; (2) {\bf Food recommendation}~\cite{Elsweiler+SIGIR-2017,Fadhil-2018}, including ingredient identification~\cite{Chen+2016}, dietary recommendation for diabetics~\cite{Rehman+2017} and recipe popularity prediction~\cite{Sanjo+2017}; (3) {\bf Cross-modal food retrieval} for image to recipe and recipe to image queries~\cite{Carvalho+SIGIR2018_AdaMine,JinJinChen+MM2018_AMSR,JinJinChen+MM2017_SAN,Salvador+CVPR2017_JESR,Hao+CVPR2019_ACME,zhu2019r2gan,lien2020recipe,fu2020mcen}. The most relevant work is the neural network approaches to learning cross-modal joint embeddings, represented by early efforts~\cite{Food101,JinJinChen+MM2017_SAN, Rasiwasia+MM2010,Frome+2013,Feng+MM2014,Wang-2017} and most recent work, represented by JESR~\cite{Salvador+CVPR2017_JESR} and its extensions~\cite{Carvalho+SIGIR2018_AdaMine,JinJinChen+MM2018_AMSR,Hao+CVPR2019_ACME,zhu2019r2gan,fu2020mcen}.
A recent effort~\cite{Xie+CogMI2020} also provides a framework to compare some of the existing methods.
SAN~\cite{JinJinChen+MM2017_SAN} applied a stacked attention network to locate ingredient regions in the image and learn joint embedding features. JESR~\cite{Salvador+CVPR2017_JESR} introduced the Recipe1M dataset and proposed a joint embedding learning approach by combining a pairwise cosine loss with a semantic regularization constraint. A few follow-up efforts have improved JESR by replacing the cosine loss optimizer. AMSR~\cite{JinJinChen+MM2018_AMSR} improves JESR by using hierarchical attention on the recipes with a simple triplet loss. AdaMine~\cite{Carvalho+SIGIR2018_AdaMine} extends JESR by using the batch all triplet loss on both recipe and image embeddings in the joint latent space to leverage class-guided features. ACME~\cite{Hao+CVPR2019_ACME} improves the cosine optimizer of JESR by using the batch-hard triplet loss~\cite{Hermans+2017} combined with adversarial cross-modal embedding~\cite{Wang-2017}. Recipe Retrieval with GAN (R$^2$GAN)~\cite{zhu2019r2gan}, Recipe Retrieval with visual query of ingredients (Img2img+JESR)~\cite{lien2020recipe} and Modality-Consistent Embedding Network (MCEN)~\cite{fu2020mcen} are recent additions, although they cannot outperform ACME.
In comparison, JEMA is the first to provide cross-modal alignment optimizations at all three tiers: learning modality-specific embedding functions, joint embedding loss optimization, and loss regularization optimization. 

 \vspace{-0.2cm}
\section{Joint Embedding with Modality Alignments}
\noindent 
Let $\mathbf{R}$ and $\mathbf{V}$ denote the recipe domain and image domain respectively. For a set $T$ of recipe-image pairs ($\mathbf{r}_{i}$, $\mathbf{v}_i$), where a recipe $\mathbf{r}_{i}\in \mathbf{R}$ and an image $\mathbf{v}_{i}\in \mathbf{V}$ ($1\leq i\leq T$), we want to jointly learn two embedding functions, $\mathbf{E}_{V}: \mathbf{V}\rightarrow \mathbb{R}^{d}$ and $\mathbf{E}_{R}: \mathbf{R}\rightarrow \mathbb{R}^{d}$, which encode each pair of raw recipe and food image into two d-dimensional vectors in the latent representation space $\mathbb{R}^{d}$: a recipe text embedding vector ($\mathbf{E}_{R}(r_i)$) and a visual image embedding vector ($\mathbf{E}_{V}(v_i)$) respectively, and the two embedding functions should satisfy the following condition: For $1 \leq i, j\leq T$, the distance between a recipe $\mathbf{r}_{i}$ and an image $\mathbf{v}_j$ in the latent d-dimensional embedding space should be closer when $i=j$ and more distant when $i\neq j$. Furthermore, good embedding functions will answer cross-modal queries with high recall performance. 
To ensure efficient learning of the two embedding functions, we argue that modality alignment optimizations should be performed at all three tiers: the learning of modality-specific embedding functions:  $\mathbf{E}_{R}$ and $\mathbf{E}_{V}$, the joint embedding for modality alignment loss optimization, and the alignment loss regularization. First, we aim to optimize the recipe text embedding by using term extraction and ranking enhanced LSTM networks for learning word and sentence sequence based recipe embedding, and at the same time, optimize the image embedding by utilizing the ResNeXt-101 as the image encoder and combining it with the image category embedding obtained by wideResNet50 and word2vec~\cite{Mikolov-2013}, as shown in Figure~\ref{general_framework}. The {\bf first-tier modality alignment optimization} achieves dual goals: the key term semantics and the category semantics help to make the recipe textual embedding more aligned with the visual features of the matching food images, and they also improve the image embedding to reflect the textual features of the matching recipes. The {\bf second-tier modality alignment} optimizes the distance-based textual-visual alignment loss by utilizing the soft-margin based batch-hard triplet loss, empowered with a novel double negative sampling strategy, denoted as $DHTL_{sm}$. Recall Figure~\ref{simple_framework}, our $DHTL_{sm}$ optimizer will improve the recipe embedding by maximizing the embedding distance in $\mathbb{R}^d$ to the {\em batch hard negative image example} chosen from those negative image embeddings in the same batch. Similarly, $DHTL_{sm}$ will optimize the image embedding by maximizing the embedding distance to the {\em batch hard negative recipe example} chosen from the negative recipe text embeddings in the same batch. The {\bf third-tier modality alignment optimization} further improves the effectiveness of our $DHTL_{sm}$ optimizer for correcting modality alignment by adding two types of loss regularizations. The category-based alignment loss aims to regularize the joint embedding loss $DHTL_{sm}$ with category-based alignment constraint on both image ($L_{CA-V}$) and recipe ($L_{CA-R}$). The cross-modal discriminator based modality alignment aims to regularize the joint embedding loss $DHTL_{sm}$ by leveraging discriminator based visual-textual distribution alignment loss ($L_{DA}$) as an alternative distance-based alignment correctness measure. 
We below define the overall objective function for joint embedding loss optimization based on the second and third tier alignment optimizations: 
\begin{equation} 
L = DHTL_{sm} + \lambda_1 L_{CA-R} +  \lambda_2  L_{CA-V} + \lambda_3  L_{DA} 
\end{equation} 
\noindent  
where $\lambda_1$, $\lambda_2$ and $\lambda_3$ are trade-off hyper-parameters. We have experimentally evaluated their impacts and observed that the best results can be obtained when $\lambda_1$, $\lambda_2$ and $\lambda_3$ are all set as 0.005.

\subsection{Modality Alignment in Embedding Functions} 
\noindent 
The first tier cross-modal optimization aims to perform early alignment of textual-visual information in learning and extracting latent features from raw training inputs of recipe and image pairs.

\noindent {\bf Recipe Text Embedding Function.\/}
We first enhance the LSTM based recipe embedding by leveraging key term features learned from recipes using term extraction and ranking algorithms. Our goal is to leverage those key terms that can uniquely distinguish a given recipe from other recipes in terms of ingredients (e.g., pork), cooking utensils (e.g., pan) and actions (e.g., stir), as illustrated in the top red dashed box in Figure~\ref{general_framework}. For the recipe sequence feature, a two-stage LSTM for representation learning of cooking instructions~\cite{Salvador+CVPR2017_JESR} is employed for representation learning of cooking instructions. 
As for the ingredients, we use the bi-directional LSTM networks to analyze the word sequence of each ingredient text, such as ``two pounds weight pork loin", and learn to identify those word sequences that are ingredients with high probability, like ``pork loin". Then we perform binary logistic regression on the extracted word sequence set to produce two clusters, labeled as true or false in the context of core ingredients. Those with true labels are regarded as the ingredient entities.
To reduce errors, we re-examine those word sequences labeled as false by leveraging the set of true ingredient entities identified and collected from the whole training dataset. If a false-labeled word sequence contains any true ingredient entity, this ingredient entity will be extracted from the false word sequence, which ensures that our recipe text embedding learning does not miss any key ingredient. Figure~\ref{miss-ingr} shows an illustrative example. For the cooking utensils and actions, we resort to the part-of-speech tagging approach~\cite{loria2018textblob} to extract the nouns and verbs in the recipe text where the ingredient entities are removed in advance. 
Given a training set of recipes, ingredients such as ``olive oil'', ``salt'' are common and more frequent than other less common ingredients, which may be unique to a small percentage of recipes. We utilize a term ranking algorithm (e.g., TFIDF~\cite{Salton-1988}, TextRank~\cite{mihalcea2004_textrank} or BERT~\cite{devlin2018_bert} based approach) to capture the discrimination significance contributed by each key term to a given recipe and the visual information in its associated food image. To generate the key term feature, a word2vec model~\cite{Mikolov-2013} is trained over the recipe texts on the entire training data to obtain the term embedding for each extracted key term. By combining the word2vec vectors of all key terms weighted by their term ranking values, we get the recipe key term feature, which is integrated with bi-directional LSTM as the training data preprocessing step. Finally, during the recipe text embedding learning phase, we combine the recipe sequence patterns and the key term features through concatenation and a fully connected layer to produce the recipe embedding in the $d$-dimensional latent space $\mathbb{R}^d$. Our JEMA recipe text embedding function can capture the key terms that are discriminative with respect to a specific recipe, and better reflect the visual components in the corresponding food image.

\noindent {\bf Food Image Embedding Function.\/} In JEMA, we also optimize the image encoder used in existing methods, e.g., VGG-16~\cite{Simonyan-2014} and ResNet-50~\cite{He+CVPR2016}. We integrate the ResNeXt-101 model~\cite{xie2017aggregated} with the category-based visual feature extracted by combining the Food-101 classifier using wideResNet-50~\cite{wideResnet} and word2vec, followed by a fully-connected layer to get the final image embedding, as shown in the blue dotted box of Figure~\ref{general_framework}. This category-based enhancement is motivated by two observations. First, the neural features learned directly from the pixel grid of the image may not always be aligned well with the neural features learned from LSTM based recipe embedding. For example, some ingredients identified by LSTM networks, such as ``olive oil'', ``salt'', may not have visual correspondence in its associated image. Furthermore, the visual appearance of some ingredients, such as ``pork\_loin'' may be missed by LSTM based recipe embedding, recall Figure~\ref{miss-ingr}. Such text-visual alignment mismatch problems are common in food computing since a cooking recipe may have different visual appearances due to different ways of using ingredients and different decorations. 

 \begin{figure}
  \centering   
  \includegraphics[scale=0.3]{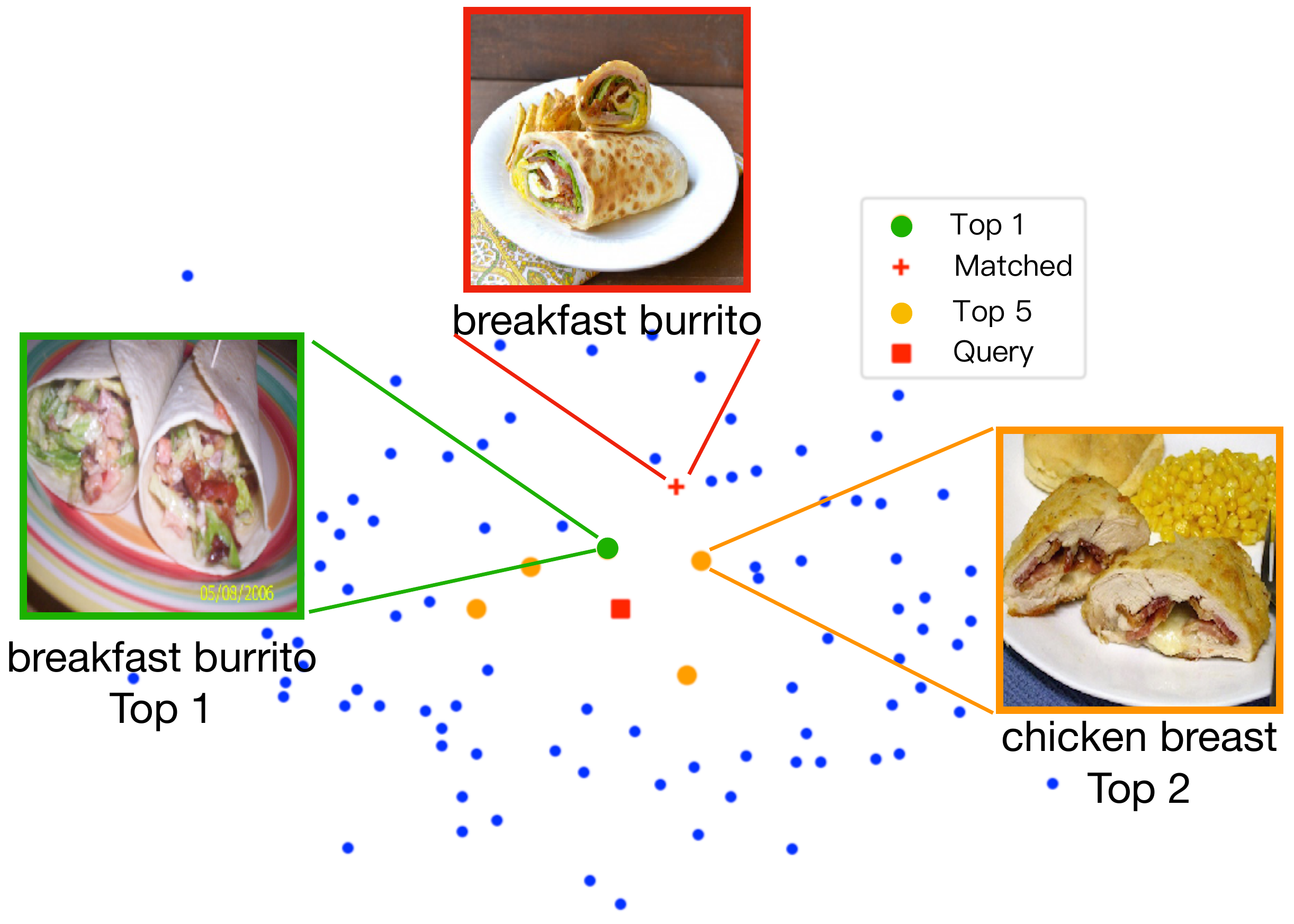}   
  \vspace{-0.4cm}
  \caption{{\footnotesize Illustration of instance-class double hard sampling strategy.}} 
  \label{negative_sample}   
    \vspace{-0.6cm}
\end{figure}   

\subsection{Joint Embedding Alignment Optimizations}
\noindent
We perform the cross-modal joint embedding on the two embeddings using a distance-based loss optimizer, which performs the second tier modality alignment optimization between the recipe and image embeddings in the $d$-dimensional latent space $\mathbb{R}^d$. In JEMA, we improve the batch hard triplet loss by introducing a double hard sampling strategy and a soft-margin function to optimize modality alignment loss.  
Triplet loss~\cite{Hermans+2017} is calculated on the triplet of training samples $(x_a, x_p, x_n)$, where $x_a$ represents a feature embedding as an anchor point in one modality and used as the ground truth to evaluate the corresponding modality embedding, $x_p$ and $x_n$ denote the positive and negative feature embeddings from the other modality. The triplet loss ensures that the positive instance in one modality should be close to the anchor point in the other modality, and the negative instance in one modality should be distant from the anchor point in the other modality. By selecting the hardest positive and negative samples for each anchor point within every batch when calculating the triplet loss, Hermans~\cite{Hermans+2017} shows that it often outperforms the {\em batch-all triplet loss}, which is based on the average distance from all negative examples to the anchor point.

\noindent {\bf Double Hard Sampling Method.\/}
In JEMA, we define two types of batch hard positive or negative examples. For instance level, given the anchor image (or recipe) embedding (i.e., $\in \mathbb{R}^d$) in each batch, there is only one positive recipe (or image) instance in the batch, which corresponds to this anchor image (or recipe), we call it the batch hardest positive example. For the rest of the recipe (or image) instances in the batch, they are considered negative examples of the anchor image (or recipe). We define the batch hardest negative recipe (or image) instance concerning this anchor as the one whose vector distance to the anchor image (or anchor recipe) is the smallest in the batch. In addition to instance level sampling, we also introduce class level sampling for computing the batch hard triplet loss. Those recipe (or image) instances that have the same category as the anchor image (or recipe) are considered as positive examples, and the rest with different categories are considered as negative examples for this anchor. We define the batch hardest positive or negative recipe (or image) example as the positive or negative example whose vector distance to the anchor image (or recipe) is the largest or smallest among all positive or negative examples in the batch respectively. 

\noindent {\bf Soft-margin based Batch Hard Triplet Loss ($DHTL_{sm}$).\/} In addition, we use the softplus function $ln(1+\exp(\gamma(\cdot+m)))$ as a smooth approximation to replace the hinge function $[m+\cdot]_+$ used in existing works~\cite{Carvalho+SIGIR2018_AdaMine,JinJinChen+MM2018_AMSR,Hao+CVPR2019_ACME}, which assumes that the distance between the anchor point and negative instance is always larger than the distance between the anchor and positive instance by a fixed margin $m$. 
Our soft-margin based approach improves the hinge with an exponential decay instead of a threshold-based hard cut-off. 
The soft-margin based batch-hard triplet loss with a double sampling strategy, denoted by $DHTL_{sm}$, is given as:
{\small
\begin{equation}  
\setlength\abovedisplayskip{3pt}
\setlength\belowdisplayskip{3pt}
\label{DHTL}  
\begin{aligned}  
DHTL_{sm} \!&=\! \sum^{N}_{i=1}ln(1\!+\!e^{\gamma({d(E_{r_i}^a, E_{v_i}^p)}\!-\!\min{d(E_{r_i}^a, E_{v_i}^n)} + m)})  \\ &+\! \sum^{N}_{i=1}ln(1\!+\!e^{\gamma( d(E_{v_i}^a, E_{r_i}^p)\!-\!\min d(E_{v_i}^a, E_{r_i}^n) + m)} )  
 \end{aligned}  
 \end{equation} 
 }
\noindent  
where $d(\cdot)$ measures the Euclidean distance between two input vectors, $N$ is the number of the different recipe-image pairs in a batch, subscripts $a$, $p$ and $n$ refer to anchor, positive and negative instances respectively, $E_{r_i}, E_{v_i}$ refer to the embeddings of the recipe and image in the $i$-th recipe-image pair respectively, $\gamma$ is the scaling factor and $m$ denotes the margin of error in the triplet loss.

\begin{figure}
  \centering   
  \includegraphics[scale=0.4]{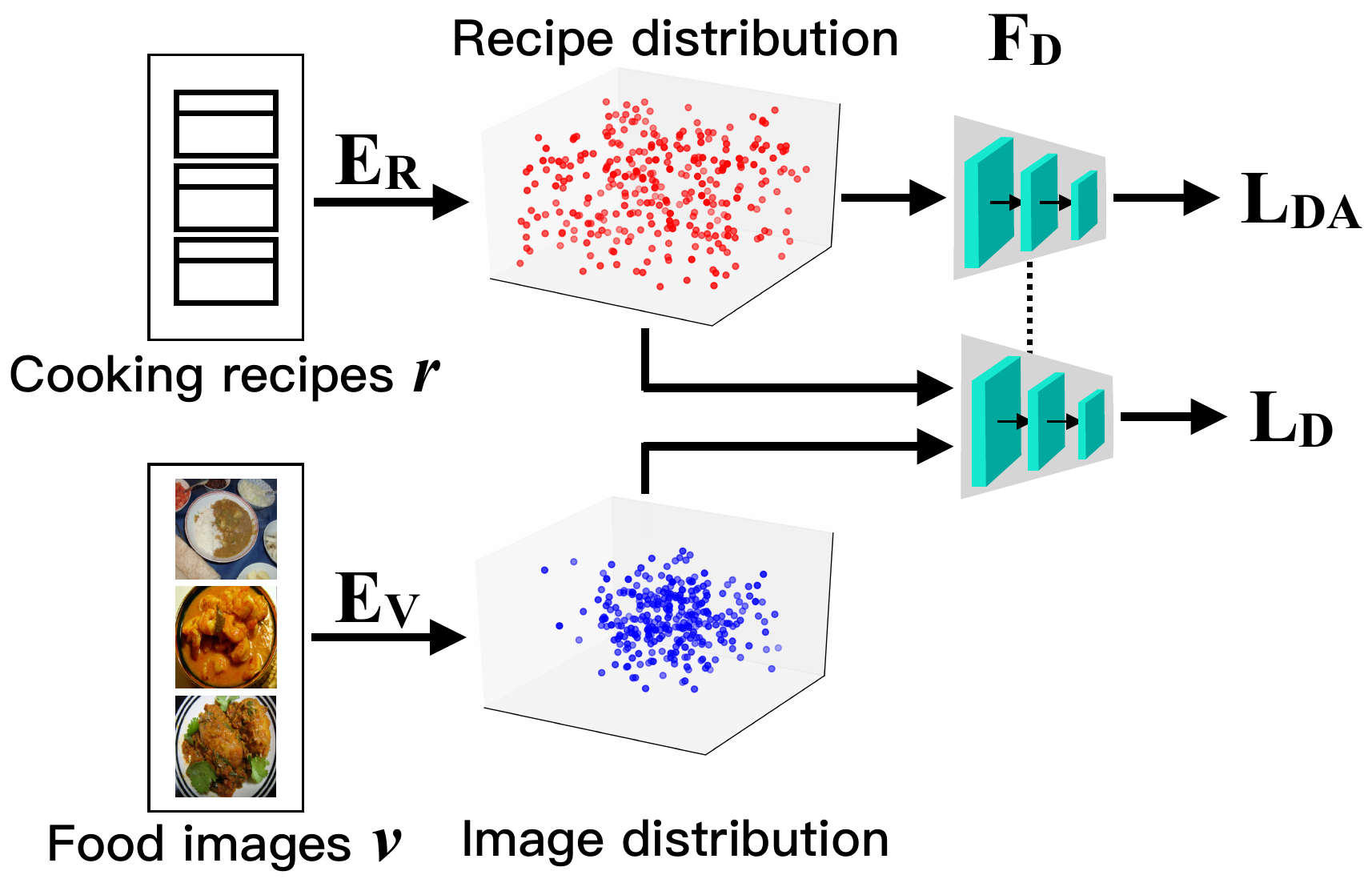}   
  \vspace{-0.2cm}
  \caption{{\footnotesize Illustration of discriminator based alignment loss regularization.} }
  \label{discriminator_nets}   
  \vspace{-0.4cm}
\end{figure}

\noindent Figure~\ref{negative_sample} illustrates the effectiveness of our instance-class double sampling strategy used in our $DHTL_{sm}$ alignment loss optimizer. 
When only using the instance level batch-hard strategy, given a recipe query (denoted as the red square), the images in the red and green boxes would be the hardest positive and negative instances respectively. However, this negative instance shares the same category ``breakfast burrito'' as the anchor recipe query. In comparison, the class level batch hard negative example would be the chicken breast image in the orange box, which is the next closest negative instance to the anchor recipe. Therefore, given a recipe query about burrito, JEMA is more likely to return the results about burrito, rather than those about chicken breast.

\subsection{ Modality Alignment Loss Regularizations} 
\noindent {\bf Category-based Alignment Loss Regularization.\/} 
The category-based loss regularization aims to reduce the cross-entropy loss between each of the $N$ modality-specific embeddings and the corresponding category from the total of $N_{c}$ categories obtained from the recipe text and the associated food image (recall Figure~\ref{general_framework}).  
JEMA assigns every recipe-image pair to a category label without using the background class. The category assignment algorithm is based on the combination of the category labels of food101~\cite{Food101}, the bigram analysis on recipe titles in the Recipe1M dataset, and the TFIDF feature on the title and recipe text. As a result, JEMA obtains 1,005 category labels, avoiding assigning background labels to a large percentage of recipe-image pairs as done in existing approaches~\cite{Salvador+CVPR2017_JESR,Carvalho+SIGIR2018_AdaMine,JinJinChen+MM2018_AMSR,Hao+CVPR2019_ACME}. 
We utilize the cross-modal category distribution alignment between the textual recipe and visual image as a regularization to our joint embedding loss optimization by $DHTL_{sm}$. The category-based loss regularization is applied to both image and recipe as follows:
{\small
\begin{equation}  
\begin{array}{c}  
L_{CA-R} = -\sum^{N}_{i=1}\sum^{N_c}_{t=1}{y^{i,t}_R \log(\hat{y}^{i,t}_R)}  
\vspace{1ex} \\  
L_{CA-V} = -\sum^{N}_{i=1}\sum^{N_c}_{t=1}{y^{i,t}_V \log(\hat{y}^{i,t}_V)}  
 \end{array}  
 \end{equation} 
}
\noindent   
where $L_{CA-R}$ is the loss of regularization on the recipe embedding, while $L_{CA-V}$ is on the image embedding. $N_c$ is the number of category labels, $N$ is the number of the different recipe-image pairs in a batch, $y^{i,t}_R$ and $\hat{y}^{i,t}_R$ are the true and estimated possibilities that the $i^{th}$ recipe embedding belongs to the $t^{th}$ category label, and similarly, $y^{i,t}_{V}$ and $\hat{y}^{i,t}_{V}$ are defined for image embedding.  

\begin{table*} 
		\center 
		\vspace{-0.2cm}
		\caption{{\small Performance comparison of our JEMA with eight existing representative methods on both 1K and 10K test set. The results of SAN, JESR,Img2img+JESR, AMSR, AdaMine, R$^2$GAN, MCEN and ACME are quoted from ~\cite{Salvador+CVPR2017_JESR,JinJinChen+MM2018_AMSR,Carvalho+SIGIR2018_AdaMine,Hao+CVPR2019_ACME,zhu2019r2gan,fu2020mcen,lien2020recipe} respectively. The symbol ``-'' indicates that the results are not available from the corresponding methods.}} 
		\label{main_results} 
		\begin{tabular}{cc|cccc|cccc} 
		\hline 
		\multirow{2} * {\tabincell{c}{Size of \\test-set}} & \multirow{2} * {Approaches}  &\multicolumn{4}{c}{Image to recipe retrieval} & \multicolumn{4}{c}{Recipe to image retrieval } \\ 
		 \cline{3-10} 
		   ~ & ~ & MedR$\downarrow$ & R@1$\uparrow$ &R@5$\uparrow$ & R@10$\uparrow$ & MedR$\downarrow$ & R@1$\uparrow$ &R@5$\uparrow$ & R@10$\uparrow$ \\   
		\hline 
		\multirow{14} *{1K}&  SAN (MMM 2017)~\cite{JinJinChen+MM2017_SAN}&16.1 & 12.5 & 31.1 & 42.3 & - & - & - & - \\ 
		~ & JESR (CVPR 2017)~\cite{Salvador+CVPR2017_JESR}  & 5.2 & 24.0 & 51.0 & 65.0 & 5.1 & 25.0 & 52.0 & 65.0 \\ 
		~ & Img2img+JESR (SIGIR 2020)~\cite{lien2020recipe}  & - & - & - & - & 5.1 & 23.9 & 51.3 & 64.1 \\ 
		~ & AMSR (MM 2018)~\cite{JinJinChen+MM2018_AMSR}  & 4.6 & 25.6 & 53.7 & 66.9 & 4.6 & 25.7 & 53.9 & 67.1 \\ 
		~ & AdaMine (SIGIR 2018)~\cite{Carvalho+SIGIR2018_AdaMine}  & 1.0 & 39.8 & 69.0 & 77.4 & 1.0 & 40.2 & 68.1 & 78.7 \\ 
		~ & R$^2$GAN (CVPR 2019) ~\cite{zhu2019r2gan} &2.0 & 39.1 & 71.0 & 81.7 & 2.0 & 40.6 & 72.6 & 83.3 \\
		~ & MCEN (CVPR 2020)~\cite{fu2020mcen} &2.0 & 48.2 & 75.8 & 83.6 & 1.9 & 48.4 & 76.1 & 83.7 \\
		~ & ACME (CVPR 2019)~\cite{Hao+CVPR2019_ACME} & 1.0 & 51.8 & 80.2 & 87.5 & 1.0 & 52.8 & 80.2 & 87.6 \\ 
		 \cline{2-10} 
		~ & \textbf{JEMA}(TextRank, ResNet-50) & 1.0 & 51.9  & 81.5 & 88.9 & 1.0  & 53.0  & 82.1  & 89.1 \\ 
		~ & \textbf{JEMA}(DistilBERT, ResNet-50) & 1.0 & 52.3  & 81.6 & 88.6 & 1.0  & 53.6  & 82.0  & 89.2 \\ 
		~ & \textbf{JEMA}(BERT, ResNet-50) & 1.0 & 52.8  & 81.7 & 89.2 & 1.0  & 53.7  & 82.3  & 89.5 \\ 
		~ & \textbf{JEMA}(RoBERTa, ResNet-50) & 1.0 & 54.6  & 83.3 & 90.4 & 1.0  & 55.2  & 83.7  & 90.7 \\ 
		~ & \textbf{JEMA}(TFIDF, ResNet-50) & 1.0 & 57.2  & 85.2 & 91.2 & 1.0  & 57.4  & 85.7  & 91.7 \\ 
		~ & \textbf{JEMA(TFIDF, ResNeXt-101)} & \textbf{1.0} & \textbf{58.1}  & \textbf{85.8} & \textbf{92.2} & \textbf{1.0}  & \textbf{58.5}  & \textbf{86.2}  & \textbf{92.3}  \\ 
		\hline 
		 \multirow{12} *{10K} & JESR (CVPR 2017)~\cite{Salvador+CVPR2017_JESR} & 41.9 & - & - & - & 39.2 & - & - & - \\ 
		~ & AMSR (MM 2018)~\cite{JinJinChen+MM2018_AMSR}   & 39.8 & 7.2 & 19.2 & 27.6 & 38.1 & 7.0 & 19.4 & 27.8 \\ 
		~ & AdaMine (SIGIR 2018)~\cite{Carvalho+SIGIR2018_AdaMine}  & 13.2 & 14.9 & 35.3 & 45.2 & 12.2 & 14.8 & 34.6 & 46.1 \\ 
		~ & R$^2$GAN (CVPR 2019) ~\cite{zhu2019r2gan} &13.9 & 13.5 & 33.5 & 44.9 & 12.6 & 14.2 & 35.0 & 46.8 \\
		~ & MCEN (CVPR 2020)~\cite{fu2020mcen} & 7.2 & 20.3 & 43.3 & 54.4 & 6.6 & 21.4 & 44.3 & 55.2 \\
		~ & ACME (CVPR 2019)~\cite{Hao+CVPR2019_ACME}& 6.7 & 22.9 & 46.8 & 57.9 & 6.0 & 24.4 & 47.9 & 59.0 \\ 
		 \cline{2-10} 
		~ & \textbf{JEMA}(TextRank, ResNet-50) & 6.2 & 21.9  & 47.2 & 59.2 & 6.0  & 22.8  & 48.1  & 59.9 \\ 
		~ & \textbf{JEMA}(DistilBERT, ResNet-50) & 6.0 & 22.7  & 48.3 & 60.1 & 6.0  & 23.7  & 48.8 & 60.5 \\ 
		~ & \textbf{JEMA}(BERT, ResNet-50) & 6.0 & 23.2  & 48.8 & 60.6 & 6.0  & 24.0  & 49.6  & 61.2 \\ 
		~ & \textbf{JEMA}(RoBERTa, ResNet-50) & 5.1 & 24.0  & 50.3 & 62.2 & 5.0  & 24.9  & 50.8  & 62.6 \\
		~ & \textbf{JEMA}(TFIDF, ResNet-50) & 4.9 & 26.3  & 53.2 & 64.9 & 4.8  & 27.0  & 53.7  & 65.3  \\ 
		~ & \textbf{JEMA(TFIDF, ResNeXt-101)} & \textbf{4.2} & \textbf{26.9}  & \textbf{54.0} & \textbf{65.6} & \textbf{4.0}  & \textbf{27.2}  & \textbf{54.4}  & \textbf{66.1}  \\ 
		\hline 
		\end{tabular} 
\end{table*} 

\begin{figure*}  [h] 
  \centering   
  \includegraphics[scale=0.23]{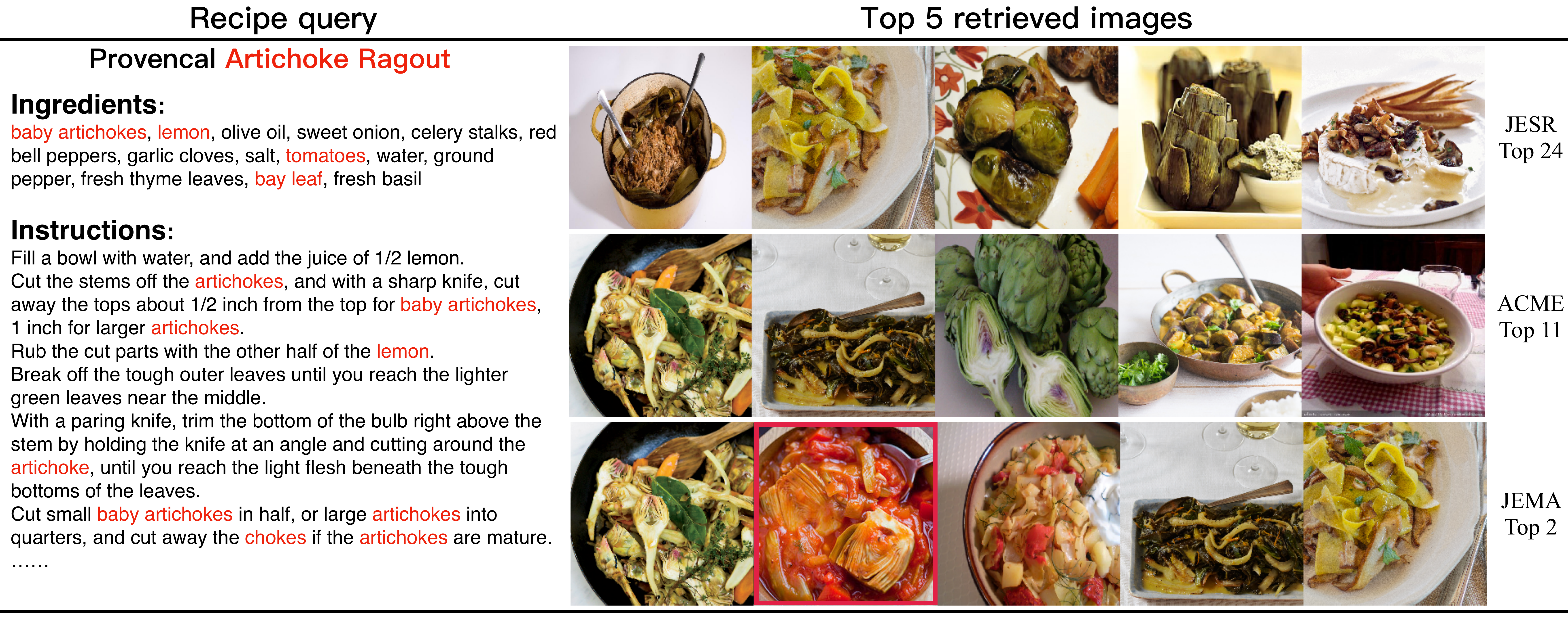}   
  \vspace{-0.2cm}
  \caption{\small{Comparing our JEMA approach with two representative methods JESR~\cite{Salvador+CVPR2017_JESR} and ACME~\cite{Hao+CVPR2019_ACME} on the recipe-to-image retrieval (1K test set). The matched images are marked in the red box. Words in red highlighted in recipe text indicate that they are selected with relatively high TFIDF value in our JEMA. We indicate the top-$k$ position where the matched image is retrieved under each of the three methods.}    }
  \label{main-r2i results}   
  \vspace{-0.2cm}
\end{figure*}

\noindent {\bf Cross-Modal Discriminator based Alignment Loss Regularization.\/}
This alignment loss regularization aims to utilize the competing strategy like those in GAN~\cite{goodfellow+2014} with the gradient penalty~\cite{gulrajani+2017} to further reduce possible errors inherent in our soft-margin based double batch hard triplet loss ($DHTL_{sm}$). To further regularize the joint embedding loss optimization, during the learning of joint embedding, we also train a discriminator model such that for each pair of matched recipe and image, given an embedding from one modality, this discriminator can tell it is the image embedding or the embedding for the matching recipe text, which is against our goal of joint embedding learning. If the trained discriminator cannot accurately discriminate the embedding of one modality from the embedding of the other modality, it indicates that our joint embedding learning is very effective and the learned distributions of embeddings for matched recipe-image pairs are too close in the common latent space $\mathbb{R}^d$ for the discriminator to tell them apart. Since the recipe and image embeddings play the same role during the discriminator learning, we select the image embedding as the target of the learned discriminator, which means receiving image embeddings, the learned discriminator would give a high confidence value while recipe embedding would result in low confidence value. The higher confidence values of recipe embeddings are obtained by our discriminator, the more aligned the distributions of recipe embeddings and image embeddings are in the joint space $\mathbb{R}^d$. The illustration of how discriminator works is showed in Figure~\ref{discriminator_nets}. We below define the loss $L_D$ for the discriminator networks, which is made up of three fully-connected layers and the cross-modal discriminator based alignment loss regularization $L_{DA}$ respectively:
{\small 
\begin{equation} 
\setlength\abovedisplayskip{3pt}
\setlength\belowdisplayskip{3pt}
\begin{aligned} 
L_D&=\sum^{B}_{i=1}[\log(F_D(E_R(r_i)))+\log(1-F_D(E_V(v_i)))\\
 &+\lambda_D(\Vert \nabla_{x_i}\log(F_D(x_i))\Vert_2-1)^2] 
 \end{aligned} 
 \end{equation} 
\begin{equation} 
L_{DA} = \sum^{B}_{i=1}{\log(1-F_D(E_R(r_i)))}  
\end{equation} 
}
\noindent 
where $F_D(\cdot)$ is the function of our trained discriminator, which outputs the confidence value in terms of how confident it is to distinguish the input embedding as the image embedding, $B$ is the number of the different recipe-image pairs in a batch, $\lambda_D$ is the trade-off parameter and set to 10 as suggested in~\cite{gulrajani+2017}, $x_i$ is a random interpolation between the $i^{th}$ recipe embedding $E_R(r_i)$ and image embedding $E_V(v_i)$.

\section{ Experiments}
\label{experiments}

 {\bf Dataset and Evaluation Metrics.}
We evaluate the effectiveness of different approaches on Recipe1M dataset~\cite{Salvador+CVPR2017_JESR}.
Experiment setup follows the literature: (i) Sample 10 unique subsets of 1,000 or 10,000 matching recipe-image pairs from the test set. (ii) Use each item in one modality as a query (e.g., an image), and rank instances in the other modality (e.g., recipes) by the Euclidean distance between the query embedding and each candidate embedding from the other modality in the test set. (iii) Calculate the median retrieval rank (MedR) and the recall percentage at top K (R@K), i.e., the percentage of queries for which the matching answer is included in the top K results (K=1,5,10). Note that even though many loss regularizations are added in our framework, these optimizations are only performed at the training stage and not used in the testing stage, only adding a little cost in the training phase and no cost in the testing.  

{\bf Implementation Details.} Word2vec model is trained using the Continuous Bag-of-Words (CBOW) architecture~\cite{word2vec-2013} on the corpus of recipe texts in the Recipe1M dataset. The dimensions of joint embedding and word2vec embedding are set as 1024 and 300 respectively. Adam optimizer~\cite{Adam-2014} is employed for model training with the initial learning rate set as $10^{-4}$ in all experiments, with the mini bath size of 100. All deep neural networks are implemented on the Pytorch platform and trained on a single Nvidia Titan X Pascal server with 12GB of memory.

{\bf Baselines for Comparison.} Eight baselines are considered: SAN~\cite{JinJinChen+MM2017_SAN}, JESR~\cite{Salvador+CVPR2017_JESR}, Img2img+JESR~\cite{lien2020recipe}, AMSR~\cite{JinJinChen+MM2018_AMSR}, AdaMine~\cite{Carvalho+SIGIR2018_AdaMine}, R$^2$GAN~\cite{zhu2019r2gan}, MCEN~\cite{fu2020mcen} and ACME~\cite{Hao+CVPR2019_ACME}.

\begin{figure}
  \centering   
  \includegraphics[scale=0.65]{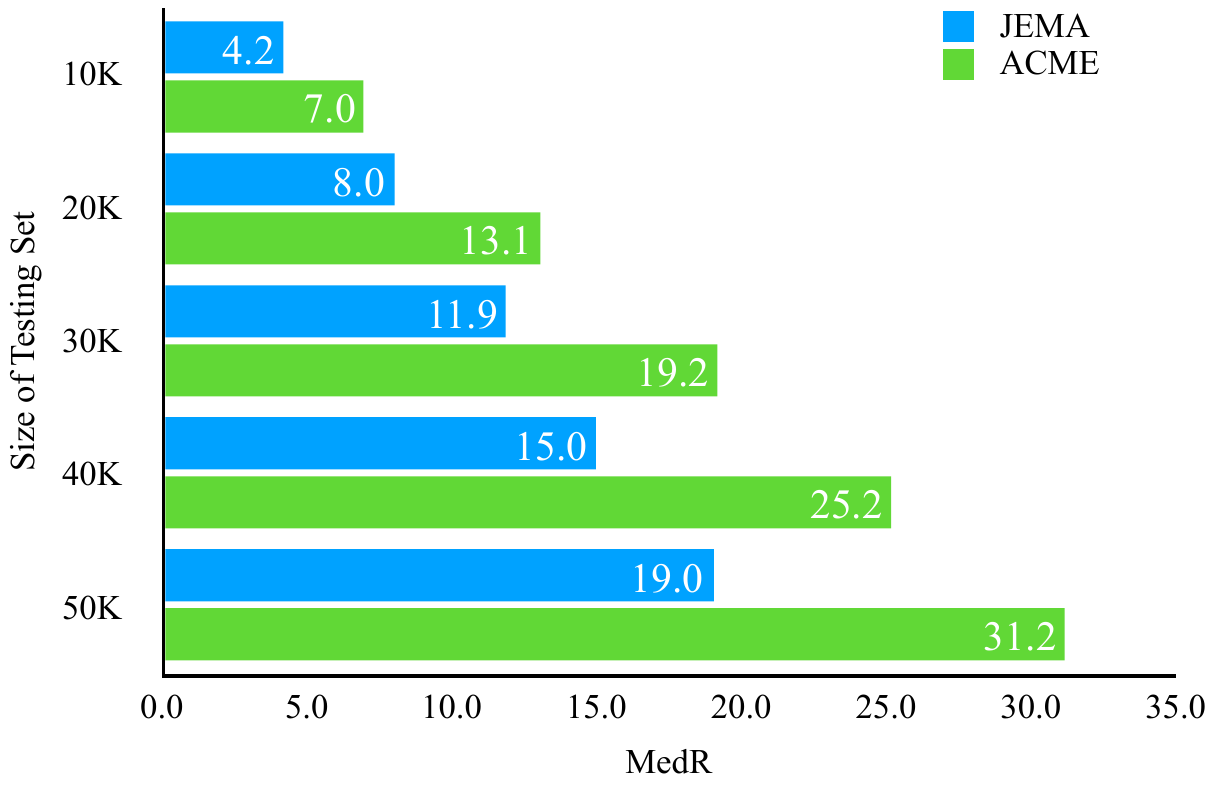}   
  \vspace{-0.2cm}
  \caption{{\small Scalability test between JEMA and ACME for image-to-recipe retrieval (For MedR, lower is better).} }
  \label{MedR}   
    \vspace{-0.2cm}
\end{figure}

\vspace{-0.2cm}
\subsection{ Cross-Modal Retrieval Performance}
\noindent
We evaluate the performance of the JEMA approach for image-to-recipe and recipe-to-image retrieval tasks against eight existing methods. Table~\ref{main_results} shows the results. We make two observations: 
(1) JEMA consistently outperforms all eight baselines with high Recall@K (K=1,5,10) for both image-to-recipe and recipe-to-image queries on 1K and 10K test data, showing the effectiveness of our three-tier textual-visual alignment optimizations for cross-modal joint embedding learning. 
(2) The attention mechanism introduced in SAN, AMSR and MCEN on top of the neural features aims to implicitly and approximately locate the key ingredients. However, the results are not effective, especially compared to JEMA, which utilizes the key terms from the key term extraction and ranking module and the image category for optimizing the text-visual modality alignments in three stages: (i) learning the two embedding functions, (ii) learning joint embedding with alignment loss optimization and (iii) alignment loss regularization.

\begin{table}
		\center
		\vspace{-0.2cm}
		\caption{\small{Evaluation of contributions of different components of the JEMA framework on the 1K test-set. MA consists of MA$_R$ and MA$_V$.}
}
		\label{ablation results}
		\begin{tabular}{c|cccc}
		\hline
		 \multirow{2} * {Component} & \multicolumn{4}{c}{Image to recipe retrieval } \\
		 \cline{2-5}
		   ~ & MedR & R@1 &R@5 & R@10  \\  
		\hline
		JEMA-b & 4.1 & 25.9 & 56.4 & 70.1 \\
		\hline
		JEMA-b+MA$_{V}$& 3.4 & 28.1 & 59.5 & 73.1  \\
		JEMA-b+MA$_R$& 3.0 & 29.4 & 60.0 & 73.4 \\	
		JEMA-b+MA & 3.0 & 30.5 & 61.6 & 75.2 \\
		\hline
		JEMA-b+MA+CA & 2.3 & 33.5 & 68.4 & 80.9 \\
		JEMA-b+MA+DA & 2.5 & 36.0 & 65.2 & 77.3 \\
		JEMA-b+MA+DHTL$_{sm}$ & 1.6 & 47.7 & 78.6 & 87.3 \\ 
		\hline
		JEMA-b+MA$_{All}$ & \textbf{1.0}  & \textbf{57.2}  & \textbf{85.2} & \textbf{91.2}\\  	
		\hline
		\end{tabular}
		\vspace{-0.2cm}
\end{table}

\begin{table*} [h]
		\center 
		\vspace{-0.2cm}
		\caption{{\small Evaluation of contributions of each component in the extracted key terms on the 10K test-set.}} 
		\label{term component} 
		\begin{tabular}{c|cccc|cccc} 
		\hline
		 \multirow{2} * {Components}  &\multicolumn{4}{c}{Image to recipe retrieval} & \multicolumn{4}{c}{Recipe to image retrieval } \\ 
		 \cline{2-9} 
		   ~ & MedR$\downarrow$ & R@1$\uparrow$ &R@5$\uparrow$ & R@10$\uparrow$ & MedR$\downarrow$ & R@1$\uparrow$ &R@5$\uparrow$ & R@10$\uparrow$ \\   
		\hline
		 ingredient & 5.0 & 25.0  & 51.4 & 63.1 & 5.0  & 26.1  & 52.0  & 63.4 \\
		 ingredient+utensil & 5.0 & 25.6  & 52.2 & 64.1 & 5.0  & 26.4  & 52.5  & 64.2 \\ 
		 ingredient+utensil+action & 5.0  & \textbf{26.3}  & \textbf{53.2} & \textbf{64.9} & \textbf{4.9}  & \textbf{27.0}  & \textbf{53.7}  & \textbf{65.3} \\ 
		\hline 
		\end{tabular} 
\end{table*}

\begin{table*} [h]
		\center 
		\vspace{-0.2cm}
		\caption{{\small Performance comparison of our JEMA using different key term filters on the 10K test set. ResNet-50 model is used here as the image encoder. The symbol ``-'' indicates that all the extracted key terms will be used to generate the key term feature. }} 
		\label{weight threshold} 
		\begin{tabular}{cc|cccc|cccc} 
		\hline
		\multirow{2} * {\tabincell{c}{Term Ranking \\Algorithm}} & \multirow{2} * {Threshold}  &\multicolumn{4}{c}{Image to recipe retrieval} & \multicolumn{4}{c}{Recipe to image retrieval } \\ 
		 \cline{3-10} 
		   ~ & ~ & MedR$\downarrow$ & R@1$\uparrow$ &R@5$\uparrow$ & R@10$\uparrow$ & MedR$\downarrow$ & R@1$\uparrow$ &R@5$\uparrow$ & R@10$\uparrow$ \\   
		\hline
		\multirow{4} *{RoBERTa} & - & 5.0 & 24.0  & 50.3 & 62.2 & 5.0  & 24.9  & 50.8  & 62.6 \\
		~ & 0.05 & 5.0 & 25.1  & 51.7 & 63.5 & 5.0  & 25.8  & 52.3  & 64.0 \\ 
		~ & 0.10 & 5.0 & 24.2  & 50.4 & 62.5 & 5.0  & 25.3  & 51.2  & 62.7 \\ 
		~ & 0.15 & 5.0 & 24.0  & 50.3 & 62.0 & 5.0  & 25.0  & 50.8  & 62.6 \\ 
		\hline 
		 \multirow{4} *{TFIDF} & - & 5.0  & \textbf{26.3}  & \textbf{53.2} & \textbf{64.9} & \textbf{4.9}  & \textbf{27.0}  & \textbf{53.7}  & \textbf{65.3}  \\ 
		~ & 0.05 & 5.0 & 26.1  & 53.1 & 64.8 & 5.0  & 26.7  & \textbf{53.7} & 65.2 \\ 
		~ & 0.10 & 5.0 & 25.3  & 51.7 & 63.4 & 5.0  & 26.1  & 52.5  & 64.1 \\ 
		~ & 0.15 & 6.0 & 22.7  & 48.0 & 59.9 & 6.0  & 23.5  & 49.1  & 60.4 \\
		\hline
		\end{tabular} 
\end{table*}

Next, we discuss the impact of different term ranking algorithms: TFIDF, TextRank and three popular BERT models (BERT~\cite{devlin2018_bert}, DistilBERT~\cite{sanh2019_distilbert} and RoBERTa~\cite{liu2019_roberta}). TFIDF and TextRank can obtain the representative terms for the recipe considering the term occurrence frequency, while BERT methods mainly focus on the semantic correlation between each term and its recipe text. The experimental results are given in Table~\ref{main_results} under JEMA sections for 1K and 10K test data. We observe that the TFIDF approach offers the best performance in JEMA. We also vary the recent CNN models in the image embedding process by changing ResNet-50 to ResNet-152~\cite{He+CVPR2016}, WideResNet-101~\cite{wideResnet}, ResNeXt-50 and ResNeXt-101~\cite{xie2017aggregated}. It turns out that using ResNeXt-101 as the image encoder can yield stable improvement for all five term ranking algorithms, and TFIDF remains to be the best in the context of JEMA. 

To further illustrate the comparison results, we provide a
visualization of recipe-to-image retrieval using an example recipe by comparing the performance of JEMA with JESR and ACME in Figure~\ref{main-r2i results}, since JESR and ACME have released their pre-trained models. 
We observe that JEMA successfully aligns the ``tomato'' ingredient in the recipe text with the red tomato component in the food image and the matched image is returned at Top 2 by JEMA. In comparison, both JESR and ACME fail to retrieve the matched image within the top 5 results, showing some problems in modality alignment between recipe text and food image. This example further illustrates the effectiveness of our three-tier alignment optimizations for improving the quality of cross-modal retrieval tasks.

\begin{figure*} 
  \centering  
  \includegraphics[scale=0.23]{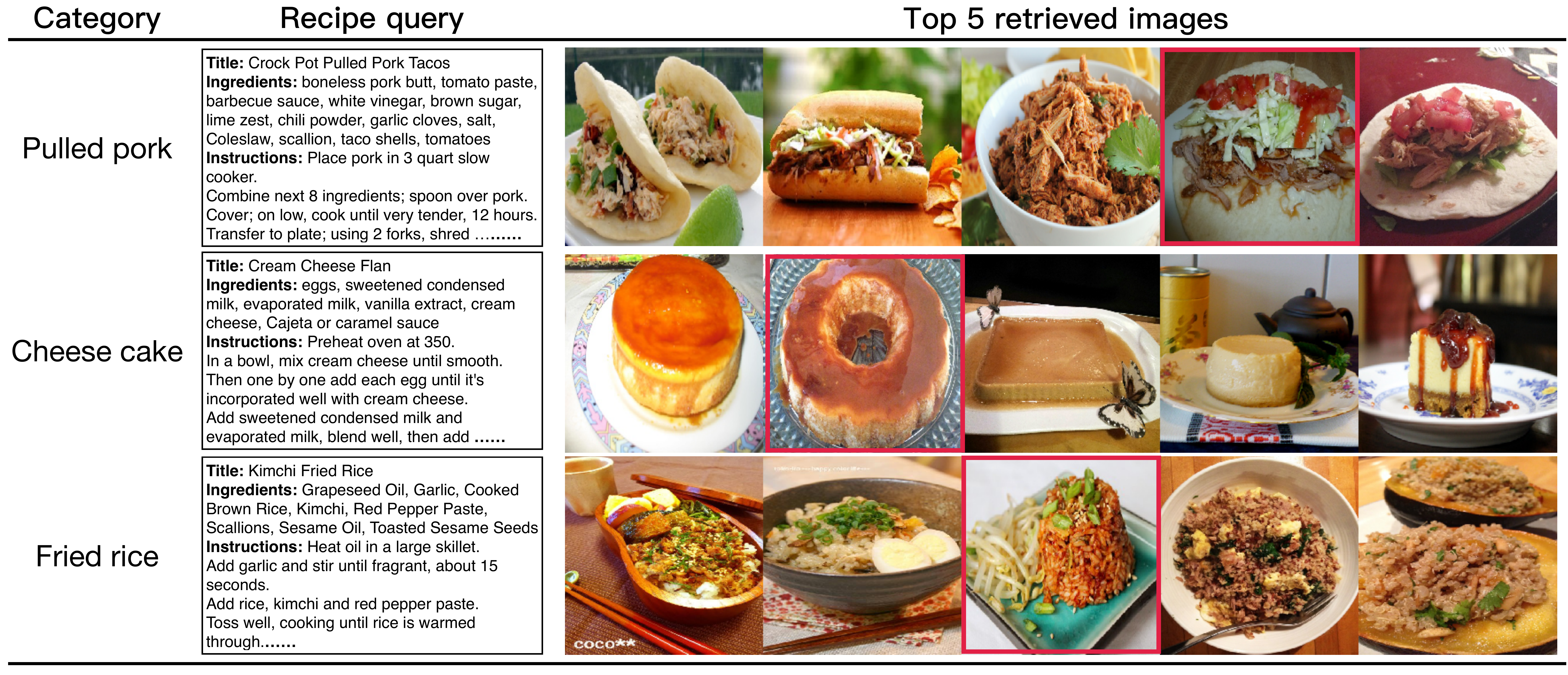}  
  \caption{{\small The results of recipe-to-image retrieval by our JEMA approach (10K test set). The matched images are boxed in red.}   }
  \label{JEMA-r2i results}  
\end{figure*}

\begin{figure*}  
  \centering  
  \includegraphics[scale=0.23]{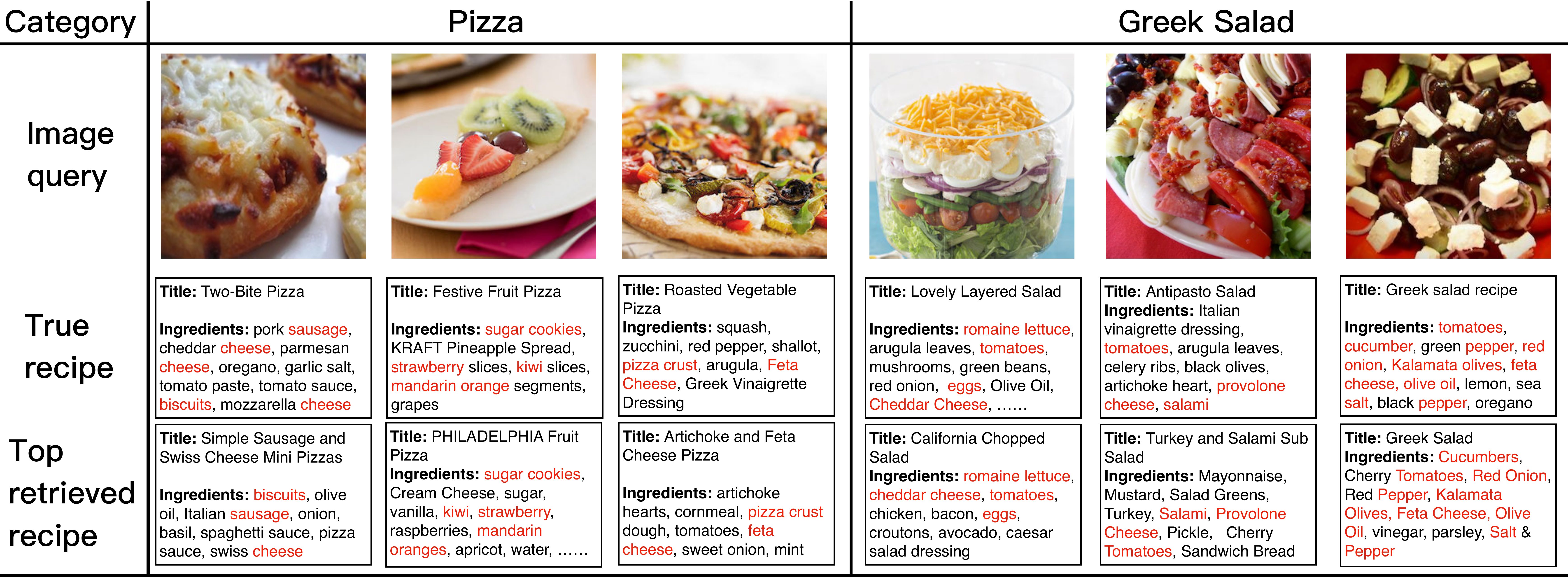}  
  \caption{{\small The results of image-to-recipe retrieval by our JEMA approach (10K test set). The common or similar ingredients in the true recipe and top retrieved recipe are highlighted in red.}   }
  \label{JEMA-i2r results}  
  \vspace{-0.1cm}
\end{figure*}

{\bf Scalability.} To investigate the robustness of JEMA against large datasets beyond 10K, we further compare its MedR performance against the state-of-the-art approach ACME. Figure~\ref{MedR} shows the results for image-to-recipe retrieval. The gap between our JEMA and ACME becomes larger as the testing set size increases. On the 50K testing set, which is almost equivalent to the original testing set~\cite{Salvador+CVPR2017_JESR}, JEMA successfully ranks the ground-truth recipes by over 12 positions ahead of ACME on average. Similar results are also observed for the recipe-to-image retrieval, where JEMA outperforms ACME by over 11 positions on average for the 50K dataset.

\subsection{ Ablation Study}
\noindent 
Ablation studies are conducted to evaluate the contributions of each core component in our JEMA approach. For a fair comparison with the baselines, ResNet-50 model is used in the image embedding and TFIDF approach is the default choice. We use \textbf{JEMA-b} to denote the joint embedding with batch-all triplet loss and without any recipe-image alignment techniques. We incrementally add one component at a time. First, we analyze the gains from employing the textual-visual alignment optimization to the two modality-specific embeddings, \textbf{MA$_V$} and \textbf{MA$_R$}. Also, we use \textbf{MA} to represent that the textual-visual alignment optimizations on both recipe and image embeddings are employed. Then we add the category-based alignment loss regularizations \textbf{CA} on both recipe and image embeddings, and add the cross-modal discriminator-based alignment loss regularization \textbf{DA}. Next, we replace the batch-all triplet loss in \textbf{JEMA-b} with our instance-class double sampling based batch-hard triplet loss \textbf{DHTL$_{sm}$}. Finally, we use \textbf{MA$_{All}$} to denote JEMA with all three-tier modality alignment optimizations. Table~\ref{ablation results} reports the contributions of these components on the image-to-recipe retrieval. It shows that every proposed component positively contributes towards improving the cross-modal alignment between recipe text and food image and the overall performance.

\subsection{ Effect of Term Extraction and Ranking}
The extracted key terms during the data preprocessing of the recipe text can be roughly divided into three types: ingredients, cooking utensils and actions. In this set of experiments, we want to evaluate the contributions of applying term extraction and ranking on each of three types of key terms on the overall performance of JEMA. 

{\bf Evaluation of Each Key Term Component.\/} For comparison, we use the TFIDF approach as the term ranking algorithm and uses ResNet-50 to generate image embedding. First, we set our JEMA approach such that it only extracts the ingredients. Then we incrementally add the other two components: the cooking utensils and the cooking actions, one at a time. Table~\ref{term component} reports the results. It shows that every proposed component positively contributes towards boosting the overall performance of cross-modal retrieval. 

{\bf Evaluation of the Key Term Filter.\/} 
In order to capture the level of discrimination significance contributed by each key term to its recipe text, in JEMA, we assign different weight to each key term extracted from the recipe text by several term ranking algorithms (i.e. TextRank, BERT based approach and TFIDF). The more discriminative the key term is for its recipe, the higher weight it will get. Since the terms with very low weights might be less useful or even harmful to the quality of the learned recipe embedding, in this set of experiments, we compare JEMA without rank-threshold based filer to denote that all key terms extracted and ranked will be utilized, which is the default, with the one using a threshold-based filter. The terms with weights lower than a threshold $t$ are removed when generating the recipe key term feature. Recall the experimental results of using different term ranking algorithms in Table~\ref{main_results}, which shows that using the RoBERTa model and TFIDF approach can get better performance. Therefore, we compare the results of generating the recipe key term features by keeping all extracted key terms with the results of using the key term filter to remove those key terms with low weight, in terms of their impacts on improving the quality of the recipe embedding learning based on the RoBERTa model and the TFIDF approach. Table~\ref{weight threshold} reports the results. Although using the key term filter on the RoBERTa based term ranking algorithm can boost the performance when the weight threshold is set to 0.05, for TFIDF, no-filter offers consistently better cross-modal retrieval performance compared to different threshold settings, showing another benefit of the TFIDF based term ranking algorithm to the performance stability of JEMA.

\subsection{ Cross-Modal Visualization Results}
\noindent We provide some visualization results of JEMA for both recipe-to-image and image-to-recipe retrieval tasks on the 10K dataset in Figure~\ref{JEMA-r2i results} and Figure~\ref{JEMA-i2r results} respectively. 
Figure~\ref{JEMA-r2i results} visualizes the results of retrieving the top 5 images using three different recipe queries. In all cases, most of the retrieved images share similar ingredients to the ground truth image. In the first example, all the retrieved images contain the visual component of pulled pork, as suggested by the category label. Also the top~1 and top~5 retrieved images both share the taco component with the ground truth image. In the second example, all top~5 results can be recognized as cheese cake and visually similar to the ground truth image. All retrieved images in the third example are visually similar and all are about fried rice with scallions. 
Figure~\ref{JEMA-i2r results} shows the ground truth recipes and the top retrieved recipes based on 6 different image queries under two recipe categories: \textit{pizza} and \textit{greek salad}. In each category, we list three image queries. We observe that almost all the significant ingredients of the true recipes and visual components in the image queries also appear in the top retrieved recipes. The results in Figure~\ref{JEMA-r2i results} and Figure~\ref{JEMA-i2r results} indicate that the joint embeddings learned by our JEMA approach are effective in aligning the textual and visual features and boosting the cross-modal retrieval performance.

   \vspace{-0.1cm}
\section{ Conclusions}
\label{conclusion} 
\noindent We have presented JEMA, a three-tier modality alignment optimization approach to learning text-image cross-modal joint embedding for cross-modal retrieval of cooking recipes and food images. This paper makes three original contributions. First, we integrate the term extraction and ranking with recipe embedding to make the textual embedding more aligned with the visual features of images. We also incorporate the image category semantics into the image embedding to better reflect the textual features in recipes. Second, we effectively optimize the textual-visual distance alignment in the joint embedding loss optimization by introducing a double sampling based batch hard triplet loss with soft-margin optimization. Third, we further reduce the joint embedding alignment loss by integrating the category-based and cross-modal discriminator based alignment loss regularizations on both recipe and image embeddings. Extensive experiments on Recipe1M benchmark dataset show that by combining the three-tier modality alignment optimizations, JEMA can effectively boost the performance of cross-modal joint embedding learning and outperform the existing representative methods for image-to-recipe and recipe-to-image retrieval tasks. 

\begin{acks}
This work is partially supported by the USA National Science Foundation under Grants NSF 2038029, 1564097, and an IBM faculty award. The first author has performed this work as a two-year visiting PhD student at Georgia Institute of Technology (2019-2021), under the support from China Scholarship Council (CSC) and Wuhan University of Technology.
\end{acks}

\bibliographystyle{ACM-Reference-Format}
\bibliography{CIKM-camera-ready}


\begin{thebibliography}{41}


\ifx \showCODEN    \undefined \def \showCODEN     #1{\unskip}     \fi
\ifx \showDOI      \undefined \def \showDOI       #1{#1}\fi
\ifx \showISBNx    \undefined \def \showISBNx     #1{\unskip}     \fi
\ifx \showISBNxiii \undefined \def \showISBNxiii  #1{\unskip}     \fi
\ifx \showISSN     \undefined \def \showISSN      #1{\unskip}     \fi
\ifx \showLCCN     \undefined \def \showLCCN      #1{\unskip}     \fi
\ifx \shownote     \undefined \def \shownote      #1{#1}          \fi
\ifx \showarticletitle \undefined \def \showarticletitle #1{#1}   \fi
\ifx \showURL      \undefined \def \showURL       {\relax}        \fi
\providecommand\bibfield[2]{#2}
\providecommand\bibinfo[2]{#2}
\providecommand\natexlab[1]{#1}
\providecommand\showeprint[2][]{arXiv:#2}

\bibitem[\protect\citeauthoryear{Bossard, Guillaumin, and Van~Gool}{Bossard
  et~al\mbox{.}}{2014}]%
        {Food101}
\bibfield{author}{\bibinfo{person}{Lukas Bossard}, \bibinfo{person}{Matthieu
  Guillaumin}, {and} \bibinfo{person}{Luc Van~Gool}.}
  \bibinfo{year}{2014}\natexlab{}.
\newblock \showarticletitle{Food-101--mining discriminative components with
  random forests}. In \bibinfo{booktitle}{\emph{European conference on computer
  vision}}. Springer, \bibinfo{pages}{446--461}.
\newblock


\bibitem[\protect\citeauthoryear{Carvalho, Cad{\`e}ne, Picard, Soulier, Thome,
  and Cord}{Carvalho et~al\mbox{.}}{2018}]%
        {Carvalho+SIGIR2018_AdaMine}
\bibfield{author}{\bibinfo{person}{Micael Carvalho}, \bibinfo{person}{R{\'e}mi
  Cad{\`e}ne}, \bibinfo{person}{David Picard}, \bibinfo{person}{Laure Soulier},
  \bibinfo{person}{Nicolas Thome}, {and} \bibinfo{person}{Matthieu Cord}.}
  \bibinfo{year}{2018}\natexlab{}.
\newblock \showarticletitle{Cross-modal retrieval in the cooking context:
  Learning semantic text-image embeddings}. In \bibinfo{booktitle}{\emph{The
  41st International ACM SIGIR Conference on Research \& Development in
  Information Retrieval}}. \bibinfo{pages}{35--44}.
\newblock


\bibitem[\protect\citeauthoryear{Chen and Ngo}{Chen and Ngo}{2016}]%
        {Chen+2016}
\bibfield{author}{\bibinfo{person}{Jingjing Chen} {and}
  \bibinfo{person}{Chong-Wah Ngo}.} \bibinfo{year}{2016}\natexlab{}.
\newblock \showarticletitle{Deep-based ingredient recognition for cooking
  recipe retrieval}. In \bibinfo{booktitle}{\emph{Proceedings of the 24th ACM
  international conference on Multimedia}}. \bibinfo{pages}{32--41}.
\newblock


\bibitem[\protect\citeauthoryear{Chen, Pang, and Ngo}{Chen
  et~al\mbox{.}}{2017}]%
        {JinJinChen+MM2017_SAN}
\bibfield{author}{\bibinfo{person}{Jingjing Chen}, \bibinfo{person}{Lei Pang},
  {and} \bibinfo{person}{Chong-Wah Ngo}.} \bibinfo{year}{2017}\natexlab{}.
\newblock \showarticletitle{Cross-modal recipe retrieval: How to cook this
  dish?}. In \bibinfo{booktitle}{\emph{International Conference on Multimedia
  Modeling}}. Springer, \bibinfo{pages}{588--600}.
\newblock


\bibitem[\protect\citeauthoryear{Chen, Ngo, Feng, and Chua}{Chen
  et~al\mbox{.}}{2018}]%
        {JinJinChen+MM2018_AMSR}
\bibfield{author}{\bibinfo{person}{Jing-Jing Chen}, \bibinfo{person}{Chong-Wah
  Ngo}, \bibinfo{person}{Fu-Li Feng}, {and} \bibinfo{person}{Tat-Seng Chua}.}
  \bibinfo{year}{2018}\natexlab{}.
\newblock \showarticletitle{Deep understanding of cooking procedure for
  cross-modal recipe retrieval}. In \bibinfo{booktitle}{\emph{Proceedings of
  the 26th ACM international conference on Multimedia}}.
  \bibinfo{pages}{1020--1028}.
\newblock


\bibitem[\protect\citeauthoryear{Devlin, Chang, Lee, and Toutanova}{Devlin
  et~al\mbox{.}}{2018}]%
        {devlin2018_bert}
\bibfield{author}{\bibinfo{person}{Jacob Devlin}, \bibinfo{person}{Ming-Wei
  Chang}, \bibinfo{person}{Kenton Lee}, {and} \bibinfo{person}{Kristina
  Toutanova}.} \bibinfo{year}{2018}\natexlab{}.
\newblock \showarticletitle{Bert: Pre-training of deep bidirectional
  transformers for language understanding}.
\newblock \bibinfo{journal}{\emph{arXiv preprint arXiv:1810.04805}}
  (\bibinfo{year}{2018}).
\newblock


\bibitem[\protect\citeauthoryear{Elsweiler, Trattner, and Harvey}{Elsweiler
  et~al\mbox{.}}{2017}]%
        {Elsweiler+SIGIR-2017}
\bibfield{author}{\bibinfo{person}{David Elsweiler}, \bibinfo{person}{Christoph
  Trattner}, {and} \bibinfo{person}{Morgan Harvey}.}
  \bibinfo{year}{2017}\natexlab{}.
\newblock \showarticletitle{Exploiting food choice biases for healthier recipe
  recommendation}. In \bibinfo{booktitle}{\emph{Proceedings of the 40th
  international acm sigir conference on research and development in information
  retrieval}}. \bibinfo{pages}{575--584}.
\newblock


\bibitem[\protect\citeauthoryear{Fadhil}{Fadhil}{2018}]%
        {Fadhil-2018}
\bibfield{author}{\bibinfo{person}{Ahmed Fadhil}.}
  \bibinfo{year}{2018}\natexlab{}.
\newblock \showarticletitle{Can a chatbot determine my diet?: Addressing
  challenges of chatbot application for meal recommendation}.
\newblock \bibinfo{journal}{\emph{arXiv preprint arXiv:1802.09100}}
  (\bibinfo{year}{2018}).
\newblock


\bibitem[\protect\citeauthoryear{Feng, Wang, and Li}{Feng
  et~al\mbox{.}}{2014}]%
        {Feng+MM2014}
\bibfield{author}{\bibinfo{person}{Fangxiang Feng}, \bibinfo{person}{Xiaojie
  Wang}, {and} \bibinfo{person}{Ruifan Li}.} \bibinfo{year}{2014}\natexlab{}.
\newblock \showarticletitle{Cross-modal retrieval with correspondence
  autoencoder}. In \bibinfo{booktitle}{\emph{Proceedings of the 22nd ACM
  international conference on Multimedia}}. \bibinfo{pages}{7--16}.
\newblock


\bibitem[\protect\citeauthoryear{Frome, Corrado, Shlens, Bengio, Dean, Ranzato,
  and Mikolov}{Frome et~al\mbox{.}}{2013}]%
        {Frome+2013}
\bibfield{author}{\bibinfo{person}{Andrea Frome}, \bibinfo{person}{Greg~S
  Corrado}, \bibinfo{person}{Jon Shlens}, \bibinfo{person}{Samy Bengio},
  \bibinfo{person}{Jeff Dean}, \bibinfo{person}{Marc'Aurelio Ranzato}, {and}
  \bibinfo{person}{Tomas Mikolov}.} \bibinfo{year}{2013}\natexlab{}.
\newblock \showarticletitle{Devise: A deep visual-semantic embedding model}. In
  \bibinfo{booktitle}{\emph{Advances in neural information processing
  systems}}. \bibinfo{pages}{2121--2129}.
\newblock


\bibitem[\protect\citeauthoryear{Fu, Wu, Liu, and Sun}{Fu
  et~al\mbox{.}}{2020}]%
        {fu2020mcen}
\bibfield{author}{\bibinfo{person}{Han Fu}, \bibinfo{person}{Rui Wu},
  \bibinfo{person}{Chenghao Liu}, {and} \bibinfo{person}{Jianling Sun}.}
  \bibinfo{year}{2020}\natexlab{}.
\newblock \showarticletitle{MCEN: Bridging Cross-Modal Gap between Cooking
  Recipes and Dish Images with Latent Variable Model}. In
  \bibinfo{booktitle}{\emph{Proceedings of the IEEE/CVF Conference on Computer
  Vision and Pattern Recognition}}. \bibinfo{pages}{14570--14580}.
\newblock


\bibitem[\protect\citeauthoryear{Goodfellow, Pouget-Abadie, Mirza, Xu,
  Warde-Farley, Ozair, Courville, and Bengio}{Goodfellow et~al\mbox{.}}{2014}]%
        {goodfellow+2014}
\bibfield{author}{\bibinfo{person}{Ian Goodfellow}, \bibinfo{person}{Jean
  Pouget-Abadie}, \bibinfo{person}{Mehdi Mirza}, \bibinfo{person}{Bing Xu},
  \bibinfo{person}{David Warde-Farley}, \bibinfo{person}{Sherjil Ozair},
  \bibinfo{person}{Aaron Courville}, {and} \bibinfo{person}{Yoshua Bengio}.}
  \bibinfo{year}{2014}\natexlab{}.
\newblock \showarticletitle{Generative adversarial nets}. In
  \bibinfo{booktitle}{\emph{Advances in neural information processing
  systems}}. \bibinfo{pages}{2672--2680}.
\newblock


\bibitem[\protect\citeauthoryear{Gulrajani, Ahmed, Arjovsky, Dumoulin, and
  Courville}{Gulrajani et~al\mbox{.}}{2017}]%
        {gulrajani+2017}
\bibfield{author}{\bibinfo{person}{Ishaan Gulrajani}, \bibinfo{person}{Faruk
  Ahmed}, \bibinfo{person}{Martin Arjovsky}, \bibinfo{person}{Vincent
  Dumoulin}, {and} \bibinfo{person}{Aaron~C Courville}.}
  \bibinfo{year}{2017}\natexlab{}.
\newblock \showarticletitle{Improved training of wasserstein gans}. In
  \bibinfo{booktitle}{\emph{Advances in neural information processing
  systems}}. \bibinfo{pages}{5767--5777}.
\newblock


\bibitem[\protect\citeauthoryear{He, Zhang, Ren, and Sun}{He
  et~al\mbox{.}}{2016}]%
        {He+CVPR2016}
\bibfield{author}{\bibinfo{person}{Kaiming He}, \bibinfo{person}{Xiangyu
  Zhang}, \bibinfo{person}{Shaoqing Ren}, {and} \bibinfo{person}{Jian Sun}.}
  \bibinfo{year}{2016}\natexlab{}.
\newblock \showarticletitle{Deep residual learning for image recognition}. In
  \bibinfo{booktitle}{\emph{Proceedings of the IEEE conference on computer
  vision and pattern recognition}}. \bibinfo{pages}{770--778}.
\newblock


\bibitem[\protect\citeauthoryear{Hermans, Beyer, and Leibe}{Hermans
  et~al\mbox{.}}{2017}]%
        {Hermans+2017}
\bibfield{author}{\bibinfo{person}{Alexander Hermans}, \bibinfo{person}{Lucas
  Beyer}, {and} \bibinfo{person}{Bastian Leibe}.}
  \bibinfo{year}{2017}\natexlab{}.
\newblock \showarticletitle{In defense of the triplet loss for person
  re-identification}.
\newblock \bibinfo{journal}{\emph{arXiv preprint arXiv:1703.07737}}
  (\bibinfo{year}{2017}).
\newblock


\bibitem[\protect\citeauthoryear{Jeon, Lavrenko, and Manmatha}{Jeon
  et~al\mbox{.}}{2003}]%
        {Jeno+SIGIR-2003}
\bibfield{author}{\bibinfo{person}{Jiwoon Jeon}, \bibinfo{person}{Victor
  Lavrenko}, {and} \bibinfo{person}{Raghavan Manmatha}.}
  \bibinfo{year}{2003}\natexlab{}.
\newblock \showarticletitle{Automatic image annotation and retrieval using
  cross-media relevance models}. In \bibinfo{booktitle}{\emph{Proceedings of
  the 26th annual international ACM SIGIR conference on Research and
  development in informaion retrieval}}. \bibinfo{pages}{119--126}.
\newblock


\bibitem[\protect\citeauthoryear{Joutou and Yanai}{Joutou and Yanai}{2009}]%
        {Joutou+ICIP2009}
\bibfield{author}{\bibinfo{person}{Taichi Joutou} {and} \bibinfo{person}{Keiji
  Yanai}.} \bibinfo{year}{2009}\natexlab{}.
\newblock \showarticletitle{A food image recognition system with multiple
  kernel learning}. In \bibinfo{booktitle}{\emph{2009 16th IEEE International
  Conference on Image Processing (ICIP)}}. IEEE, \bibinfo{pages}{285--288}.
\newblock


\bibitem[\protect\citeauthoryear{Kawano and Yanai}{Kawano and Yanai}{2014}]%
        {Kawano+2014}
\bibfield{author}{\bibinfo{person}{Yoshiyuki Kawano} {and}
  \bibinfo{person}{Keiji Yanai}.} \bibinfo{year}{2014}\natexlab{}.
\newblock \showarticletitle{Food image recognition with deep convolutional
  features}. In \bibinfo{booktitle}{\emph{Proceedings of the 2014 ACM
  International Joint Conference on Pervasive and Ubiquitous Computing: Adjunct
  Publication}}. \bibinfo{pages}{589--593}.
\newblock


\bibitem[\protect\citeauthoryear{Kingma and Ba}{Kingma and Ba}{2014}]%
        {Adam-2014}
\bibfield{author}{\bibinfo{person}{Diederik~P Kingma} {and}
  \bibinfo{person}{Jimmy Ba}.} \bibinfo{year}{2014}\natexlab{}.
\newblock \showarticletitle{Adam: A method for stochastic optimization}.
\newblock \bibinfo{journal}{\emph{arXiv preprint arXiv:1412.6980}}
  (\bibinfo{year}{2014}).
\newblock


\bibitem[\protect\citeauthoryear{Lien, Zamani, and Croft}{Lien
  et~al\mbox{.}}{2020}]%
        {lien2020recipe}
\bibfield{author}{\bibinfo{person}{Yen-Chieh Lien}, \bibinfo{person}{Hamed
  Zamani}, {and} \bibinfo{person}{W~Bruce Croft}.}
  \bibinfo{year}{2020}\natexlab{}.
\newblock \showarticletitle{Recipe Retrieval with Visual Query of Ingredients}.
  In \bibinfo{booktitle}{\emph{Proceedings of the 43rd International ACM SIGIR
  Conference on Research and Development in Information Retrieval}}.
  \bibinfo{pages}{1565--1568}.
\newblock


\bibitem[\protect\citeauthoryear{Liu, Ott, Goyal, Du, Joshi, Chen, Levy, Lewis,
  Zettlemoyer, and Stoyanov}{Liu et~al\mbox{.}}{2019}]%
        {liu2019_roberta}
\bibfield{author}{\bibinfo{person}{Yinhan Liu}, \bibinfo{person}{Myle Ott},
  \bibinfo{person}{Naman Goyal}, \bibinfo{person}{Jingfei Du},
  \bibinfo{person}{Mandar Joshi}, \bibinfo{person}{Danqi Chen},
  \bibinfo{person}{Omer Levy}, \bibinfo{person}{Mike Lewis},
  \bibinfo{person}{Luke Zettlemoyer}, {and} \bibinfo{person}{Veselin
  Stoyanov}.} \bibinfo{year}{2019}\natexlab{}.
\newblock \showarticletitle{Roberta: A robustly optimized bert pretraining
  approach}.
\newblock \bibinfo{journal}{\emph{arXiv preprint arXiv:1907.11692}}
  (\bibinfo{year}{2019}).
\newblock


\bibitem[\protect\citeauthoryear{Loria}{Loria}{2018}]%
        {loria2018textblob}
\bibfield{author}{\bibinfo{person}{Steven Loria}.}
  \bibinfo{year}{2018}\natexlab{}.
\newblock \showarticletitle{textblob Documentation}.
\newblock \bibinfo{journal}{\emph{Release 0.15}}  \bibinfo{volume}{2}
  (\bibinfo{year}{2018}).
\newblock


\bibitem[\protect\citeauthoryear{Mihalcea and Tarau}{Mihalcea and
  Tarau}{2004}]%
        {mihalcea2004_textrank}
\bibfield{author}{\bibinfo{person}{Rada Mihalcea} {and} \bibinfo{person}{Paul
  Tarau}.} \bibinfo{year}{2004}\natexlab{}.
\newblock \showarticletitle{Textrank: Bringing order into text}. In
  \bibinfo{booktitle}{\emph{Proceedings of the 2004 conference on empirical
  methods in natural language processing}}. \bibinfo{pages}{404--411}.
\newblock


\bibitem[\protect\citeauthoryear{Mikolov, Chen, Corrado, and Dean}{Mikolov
  et~al\mbox{.}}{2013a}]%
        {Mikolov-2013}
\bibfield{author}{\bibinfo{person}{Tomas Mikolov}, \bibinfo{person}{Kai Chen},
  \bibinfo{person}{Greg Corrado}, {and} \bibinfo{person}{Jeffrey Dean}.}
  \bibinfo{year}{2013}\natexlab{a}.
\newblock \showarticletitle{Efficient estimation of word representations in
  vector space}.
\newblock \bibinfo{journal}{\emph{arXiv preprint arXiv:1301.3781}}
  (\bibinfo{year}{2013}).
\newblock


\bibitem[\protect\citeauthoryear{Mikolov, Sutskever, Chen, Corrado, and
  Dean}{Mikolov et~al\mbox{.}}{2013b}]%
        {word2vec-2013}
\bibfield{author}{\bibinfo{person}{Tomas Mikolov}, \bibinfo{person}{Ilya
  Sutskever}, \bibinfo{person}{Kai Chen}, \bibinfo{person}{Greg~S Corrado},
  {and} \bibinfo{person}{Jeff Dean}.} \bibinfo{year}{2013}\natexlab{b}.
\newblock \showarticletitle{Distributed representations of words and phrases
  and their compositionality}. In \bibinfo{booktitle}{\emph{Advances in neural
  information processing systems}}. \bibinfo{pages}{3111--3119}.
\newblock


\bibitem[\protect\citeauthoryear{Rasiwasia, Costa~Pereira, Coviello, Doyle,
  Lanckriet, Levy, and Vasconcelos}{Rasiwasia et~al\mbox{.}}{2010}]%
        {Rasiwasia+MM2010}
\bibfield{author}{\bibinfo{person}{Nikhil Rasiwasia}, \bibinfo{person}{Jose
  Costa~Pereira}, \bibinfo{person}{Emanuele Coviello}, \bibinfo{person}{Gabriel
  Doyle}, \bibinfo{person}{Gert~RG Lanckriet}, \bibinfo{person}{Roger Levy},
  {and} \bibinfo{person}{Nuno Vasconcelos}.} \bibinfo{year}{2010}\natexlab{}.
\newblock \showarticletitle{A new approach to cross-modal multimedia
  retrieval}. In \bibinfo{booktitle}{\emph{Proceedings of the 18th ACM
  international conference on Multimedia}}. \bibinfo{pages}{251--260}.
\newblock


\bibitem[\protect\citeauthoryear{Rehman, Khalid, Bilal, Madani,
  et~al\mbox{.}}{Rehman et~al\mbox{.}}{2017}]%
        {Rehman+2017}
\bibfield{author}{\bibinfo{person}{Faisal Rehman}, \bibinfo{person}{Osman
  Khalid}, \bibinfo{person}{Kashif Bilal}, \bibinfo{person}{Sajjad~A Madani},
  {et~al\mbox{.}}} \bibinfo{year}{2017}\natexlab{}.
\newblock \showarticletitle{Diet-Right: A Smart Food Recommendation System.}
\newblock \bibinfo{journal}{\emph{KSII Transactions on Internet \& Information
  Systems}} \bibinfo{volume}{11}, \bibinfo{number}{6} (\bibinfo{year}{2017}).
\newblock


\bibitem[\protect\citeauthoryear{Salton and Buckley}{Salton and
  Buckley}{1988}]%
        {Salton-1988}
\bibfield{author}{\bibinfo{person}{Gerard Salton} {and}
  \bibinfo{person}{Christopher Buckley}.} \bibinfo{year}{1988}\natexlab{}.
\newblock \showarticletitle{Term-weighting approaches in automatic text
  retrieval}.
\newblock \bibinfo{journal}{\emph{Information processing \& management}}
  \bibinfo{volume}{24}, \bibinfo{number}{5} (\bibinfo{year}{1988}),
  \bibinfo{pages}{513--523}.
\newblock


\bibitem[\protect\citeauthoryear{Salvador, Hynes, Aytar, Marin, Ofli, Weber,
  and Torralba}{Salvador et~al\mbox{.}}{2017}]%
        {Salvador+CVPR2017_JESR}
\bibfield{author}{\bibinfo{person}{Amaia Salvador}, \bibinfo{person}{Nicholas
  Hynes}, \bibinfo{person}{Yusuf Aytar}, \bibinfo{person}{Javier Marin},
  \bibinfo{person}{Ferda Ofli}, \bibinfo{person}{Ingmar Weber}, {and}
  \bibinfo{person}{Antonio Torralba}.} \bibinfo{year}{2017}\natexlab{}.
\newblock \showarticletitle{Learning cross-modal embeddings for cooking recipes
  and food images}. In \bibinfo{booktitle}{\emph{Proceedings of the IEEE
  conference on computer vision and pattern recognition}}.
  \bibinfo{pages}{3020--3028}.
\newblock


\bibitem[\protect\citeauthoryear{Sanh, Debut, Chaumond, and Wolf}{Sanh
  et~al\mbox{.}}{2019}]%
        {sanh2019_distilbert}
\bibfield{author}{\bibinfo{person}{Victor Sanh}, \bibinfo{person}{Lysandre
  Debut}, \bibinfo{person}{Julien Chaumond}, {and} \bibinfo{person}{Thomas
  Wolf}.} \bibinfo{year}{2019}\natexlab{}.
\newblock \showarticletitle{DistilBERT, a distilled version of BERT: smaller,
  faster, cheaper and lighter}.
\newblock \bibinfo{journal}{\emph{arXiv preprint arXiv:1910.01108}}
  (\bibinfo{year}{2019}).
\newblock


\bibitem[\protect\citeauthoryear{Sanjo and Katsurai}{Sanjo and
  Katsurai}{2017}]%
        {Sanjo+2017}
\bibfield{author}{\bibinfo{person}{Satoshi Sanjo} {and} \bibinfo{person}{Marie
  Katsurai}.} \bibinfo{year}{2017}\natexlab{}.
\newblock \showarticletitle{Recipe popularity prediction with deep
  visual-semantic fusion}. In \bibinfo{booktitle}{\emph{Proceedings of the 2017
  ACM on Conference on Information and Knowledge Management}}.
  \bibinfo{pages}{2279--2282}.
\newblock


\bibitem[\protect\citeauthoryear{Simonyan and Zisserman}{Simonyan and
  Zisserman}{2014}]%
        {Simonyan-2014}
\bibfield{author}{\bibinfo{person}{Karen Simonyan} {and}
  \bibinfo{person}{Andrew Zisserman}.} \bibinfo{year}{2014}\natexlab{}.
\newblock \showarticletitle{Very deep convolutional networks for large-scale
  image recognition}.
\newblock \bibinfo{journal}{\emph{arXiv preprint arXiv:1409.1556}}
  (\bibinfo{year}{2014}).
\newblock


\bibitem[\protect\citeauthoryear{Sun, Bhowmick, Nam~Nguyen, and Bai}{Sun
  et~al\mbox{.}}{2011}]%
        {Sun-2011}
\bibfield{author}{\bibinfo{person}{Aixin Sun}, \bibinfo{person}{Sourav~S
  Bhowmick}, \bibinfo{person}{Khanh~Tran Nam~Nguyen}, {and} \bibinfo{person}{Ge
  Bai}.} \bibinfo{year}{2011}\natexlab{}.
\newblock \showarticletitle{Tag-based social image retrieval: An empirical
  evaluation}.
\newblock \bibinfo{journal}{\emph{Journal of the American Society for
  Information Science and Technology}} \bibinfo{volume}{62},
  \bibinfo{number}{12} (\bibinfo{year}{2011}), \bibinfo{pages}{2364--2381}.
\newblock


\bibitem[\protect\citeauthoryear{Trattner and Elsweiler}{Trattner and
  Elsweiler}{2017}]%
        {Tratter+WWW2017}
\bibfield{author}{\bibinfo{person}{Christoph Trattner} {and}
  \bibinfo{person}{David Elsweiler}.} \bibinfo{year}{2017}\natexlab{}.
\newblock \showarticletitle{Investigating the healthiness of internet-sourced
  recipes: implications for meal planning and recommender systems}. In
  \bibinfo{booktitle}{\emph{Proceedings of the 26th international conference on
  world wide web}}. \bibinfo{pages}{489--498}.
\newblock


\bibitem[\protect\citeauthoryear{Wang, Yang, Xu, Hanjalic, and Shen}{Wang
  et~al\mbox{.}}{2017}]%
        {Wang-2017}
\bibfield{author}{\bibinfo{person}{Bokun Wang}, \bibinfo{person}{Yang Yang},
  \bibinfo{person}{Xing Xu}, \bibinfo{person}{Alan Hanjalic}, {and}
  \bibinfo{person}{Heng~Tao Shen}.} \bibinfo{year}{2017}\natexlab{}.
\newblock \showarticletitle{Adversarial cross-modal retrieval}. In
  \bibinfo{booktitle}{\emph{Proceedings of the 25th ACM international
  conference on Multimedia}}. \bibinfo{pages}{154--162}.
\newblock


\bibitem[\protect\citeauthoryear{Wang, Sahoo, Liu, Lim, and Hoi}{Wang
  et~al\mbox{.}}{2019}]%
        {Hao+CVPR2019_ACME}
\bibfield{author}{\bibinfo{person}{Hao Wang}, \bibinfo{person}{Doyen Sahoo},
  \bibinfo{person}{Chenghao Liu}, \bibinfo{person}{Ee-peng Lim}, {and}
  \bibinfo{person}{Steven~CH Hoi}.} \bibinfo{year}{2019}\natexlab{}.
\newblock \showarticletitle{Learning cross-modal embeddings with adversarial
  networks for cooking recipes and food images}. In
  \bibinfo{booktitle}{\emph{Proceedings of the IEEE Conference on Computer
  Vision and Pattern Recognition}}. \bibinfo{pages}{11572--11581}.
\newblock


\bibitem[\protect\citeauthoryear{Xie, Girshick, Doll{\'a}r, Tu, and He}{Xie
  et~al\mbox{.}}{2017}]%
        {xie2017aggregated}
\bibfield{author}{\bibinfo{person}{Saining Xie}, \bibinfo{person}{Ross
  Girshick}, \bibinfo{person}{Piotr Doll{\'a}r}, \bibinfo{person}{Zhuowen Tu},
  {and} \bibinfo{person}{Kaiming He}.} \bibinfo{year}{2017}\natexlab{}.
\newblock \showarticletitle{Aggregated residual transformations for deep neural
  networks}. In \bibinfo{booktitle}{\emph{Proceedings of the IEEE conference on
  computer vision and pattern recognition}}. \bibinfo{pages}{1492--1500}.
\newblock


\bibitem[\protect\citeauthoryear{Xie, Liu, Wu, Li, and Zhong}{Xie
  et~al\mbox{.}}{2020}]%
        {Xie+CogMI2020}
\bibfield{author}{\bibinfo{person}{Zhongwei Xie}, \bibinfo{person}{Ling Liu},
  \bibinfo{person}{Yanzhao Wu}, \bibinfo{person}{Lin Li}, {and}
  \bibinfo{person}{Luo Zhong}.} \bibinfo{year}{2020}\natexlab{}.
\newblock \showarticletitle{Cross-Modal Joint Embedding with Diverse
  Semantics}. In \bibinfo{booktitle}{\emph{2020 IEEE First International
  Conference on Cognitive Machine Intelligence (CogMI)}}. IEEE.
\newblock


\bibitem[\protect\citeauthoryear{Yanai and Kawano}{Yanai and Kawano}{2015}]%
        {Yanai+2015}
\bibfield{author}{\bibinfo{person}{Keiji Yanai} {and}
  \bibinfo{person}{Yoshiyuki Kawano}.} \bibinfo{year}{2015}\natexlab{}.
\newblock \showarticletitle{Food image recognition using deep convolutional
  network with pre-training and fine-tuning}. In \bibinfo{booktitle}{\emph{2015
  IEEE International Conference on Multimedia \& Expo Workshops (ICMEW)}}.
  IEEE, \bibinfo{pages}{1--6}.
\newblock


\bibitem[\protect\citeauthoryear{Zagoruyko and Komodakis}{Zagoruyko and
  Komodakis}{2016}]%
        {wideResnet}
\bibfield{author}{\bibinfo{person}{Sergey Zagoruyko} {and}
  \bibinfo{person}{Nikos Komodakis}.} \bibinfo{year}{2016}\natexlab{}.
\newblock \showarticletitle{Wide residual networks}.
\newblock \bibinfo{journal}{\emph{arXiv preprint arXiv:1605.07146}}
  (\bibinfo{year}{2016}).
\newblock


\bibitem[\protect\citeauthoryear{Zhu, Ngo, Chen, and Hao}{Zhu
  et~al\mbox{.}}{2019}]%
        {zhu2019r2gan}
\bibfield{author}{\bibinfo{person}{Bin Zhu}, \bibinfo{person}{Chong-Wah Ngo},
  \bibinfo{person}{Jingjing Chen}, {and} \bibinfo{person}{Yanbin Hao}.}
  \bibinfo{year}{2019}\natexlab{}.
\newblock \showarticletitle{R2gan: Cross-modal recipe retrieval with generative
  adversarial network}. In \bibinfo{booktitle}{\emph{Proceedings of the
  IEEE/CVF Conference on Computer Vision and Pattern Recognition}}.
  \bibinfo{pages}{11477--11486}.
\newblock


\end{thebibliography}

\end{document}